%% file: aaai25.tex
\documentclass[letterpaper]{article} 
\usepackage{aaai25}  
\usepackage{times}  
\usepackage{helvet}  
\usepackage{courier}  
\usepackage[hyphens]{url}  
\usepackage{graphicx} 
\usepackage{natbib}  
\usepackage{caption} 
\usepackage{algorithm}
\usepackage{algorithmic}

\usepackage{newfloat}
\usepackage{listings}
\DeclareCaptionStyle{ruled}{labelfont=normalfont,labelsep=colon,strut=off} 
\floatstyle{ruled}
\newfloat{listing}{tb}{lst}{}
\floatname{listing}{Listing}
\usepackage{booktabs}
\usepackage{amsmath}
\usepackage{amssymb}
\usepackage{subcaption}
\usepackage{adjustbox}
\usepackage{multirow}
\usepackage{tabularx}
\usepackage{makecell}
\usepackage{float}
\usepackage{xcolor}
\usepackage{enumitem}

\usepackage[most]{tcolorbox}
\definecolor{light_green}{HTML}{b2ffb2}
\definecolor{sage}{HTML}{c3efb2}
\definecolor{light_red}{HTML}{ffb2b2}
\definecolor{light_blue}{HTML}{add8e6}
\tcbset{
    on line, 
    boxsep=1pt,left=1pt,right=1pt,top=1pt,bottom=1pt,
    colframe=white,
    colback=light_green,  
    highlight math style={enhanced}
}

\title{How Does the Spatial Distribution of Pre-training Data Affect\\Geospatial Foundation Models?}
\author{
    Mirali Purohit\equalcontrib\textsuperscript{\rm 1},
    Gedeon Muhawenayo\equalcontrib\textsuperscript{\rm 1},
    Esther Rolf\textsuperscript{\rm 2},
    Hannah Kerner\textsuperscript{\rm 1}
}

\affiliations{
    \textsuperscript{\rm 1}Arizona State University\\
    \textsuperscript{\rm 2}University of Colorado Boulder
    \\
}

\usepackage{bibentry}

\begin{document}

\maketitle

\begin{abstract}

Foundation models have made rapid advances in many domains including Earth observation, where Geospatial Foundation Models (GFMs) can help address global challenges such as climate change, agriculture, and disaster response. Previous work on GFMs focused on tailoring model architecture and pre-text tasks, and did not investigate the impact of pre-training data selection on model performance. However, recent works from other domains show that the pre-training data distribution is an important factor influencing the performance of the foundation models. With this motivation, our research explores how the \textit{geographic} distribution of pre-training data affects the performance of GFMs. We evaluated several pre-training data distributions by sampling different compositions from a global data pool. Our experiments with two GFMs on downstream tasks indicate that balanced and globally representative data compositions often outperform region-specific sampling, highlighting the importance of diversity and global coverage in pre-training data. Our results suggest that the most appropriate data sampling technique may depend on the specific GFM architecture. These findings will support the development of robust GFMs by incorporating quality pre-training data distributions, ultimately improving machine learning solutions for Earth observation.




\end{abstract}

%

\section{Introduction}
\label{sec:introduction}

\begin{figure*}[t]
\centering
\includegraphics[width=0.9\textwidth]{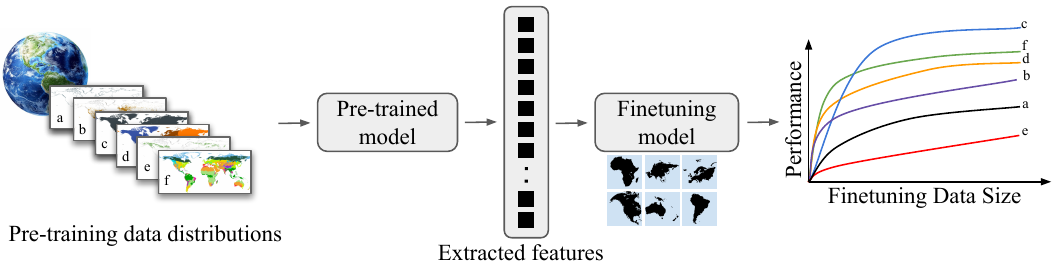}
\caption{Our experimental pipeline to measure the impact of pre-training data distributions on downstream task performance.}
\label{fig:teaser_figure}
\end{figure*}

Foundation models have proven to be powerful tools across a wide variety of fields. These models have shown remarkable potential by achieving promising results while reducing the dependencies on labeled data. Researchers are increasingly adapting foundation models in specialized domains, such as health, law, physics, and geoscience by leveraging their ability to generalize from large datasets \cite{colombo2024saullm, parmar-etal-2022-boxbart, cong2022satmae}. 

In recent years, Geospatial Foundation Models (GFMs) pre-trained on vast quantities of satellite data have been developed for Earth observation (EO) applications. These models can play a critical role in improving the performance of tasks related to climate change, agriculture, oceanography, forests, disaster response, and more \cite{jakubik2310foundation, tseng2023lightweight, klemmer2023satclip}.

There is a general perception in AI that more data will help to improve the model's performance \cite{sun2017revisiting, shi2024scaling}. Hence, most current efforts to build foundation models usually pre-train on as much data as available or computationally feasible. However, recent studies show that dataset characteristics like diversity, spatial distribution, and representivity can be more important than data quantity to build robust ML models \cite{nguyen2022quality, cole2022does}. This suggests that understanding the effects of pre-training data distribution on model performance is crucial. 

Although some studies have explored the relation between pre-training data distribution and downstream task performance in computer vision (CV) for natural images \cite{hammoud2024pretraining, fang2022data, longpre2023pretrainer}, there is a lack of studies of this type for GFMs. To build GFMs, researchers use large remote sensing datasets collected by satellites and aerial platforms. Given the massive data volume available from EO satellites (e.g., from the Sentinel-2 satellites), it is essential to select pre-training data in a way that benefits model performance but also reduces the computational cost -- e.g., by focusing on data quality instead of quantity \cite{rolf2021representationmattersassessingimportance}.

Unlike standard CV datasets, which are generally static, geospatial datasets are dynamic and available from multiple time periods. Satellite data offers the unique advantage of global coverage, allowing data to be sampled from any location on Earth. However, we don't know the answer to some critical questions, such as ``Is it crucial to have data from all continents? Is it crucial to have data evenly represented across biomes?"

As a result, existing GFMs employ different strategies to sample pre-training data, showing a \emph{lack of consensus on a unified approach to sample pre-training data} (details in \textsection \ref{sec:related_work}). \citet{roscher2023better} showed that most of the modeling focus in the geospatial domain is currently paid to the pre-training approach or model architecture.  At the same time, \emph{the composition of data chosen for pre-training is relatively understudied}, due to the assumption that training data size is the dominant factor determining performance. In particular, \citet{manas2021seasonal} is the only study that compares model performance with multiple data sampling strategies.

Motivated by these limitations, we investigated how different GFMs perform on downstream tasks when different sampling techniques are used to curate pre-training data (pipeline shown in Figure \ref{fig:teaser_figure}). We conducted experiments on two GFMs by pre-training models on five different data sampling strategies, referred as \textit{data compositions}. These strategies include random, stratified, and clustered sampling techniques (see \textsection \ref{subsec:pretraining_data_composition}). We refer to sampling techniques that ensure balanced or global data representation as balanced sampling techniques, and those grouping samples by spatial or environmental characteristics as clustered sampling. We pre-trained the models on five different data compositions and investigated how they affect the downstream task performance disaggregated by continent (Figure \ref{fig:teaser_figure}). We compared these results with a baseline of no pre-training. Our main contributions are:


\begin{itemize}[noitemsep]
    \item We investigated the effect of different pre-training data compositions on downstream tasks for GFMs. To the best of the authors' knowledge, this is the first study to explore the impact of pre-training data distribution on downstream task performance for GFMs.
    \item Our experimental results show that all balanced sampling techniques yield approximately equal performance on downstream tasks, and usually outperform models with clustered pre-training data or no pre-training.
\end{itemize}

Overall, our observations highlight the importance of balanced and representative data sampling strategies in pre-training GFMs. We believe our findings will contribute to the development of robust GFMs by incorporating quality pre-training data that benefit downstream task performance.

\section{Related Work}
\label{sec:related_work}

The relationship between pre-training data distribution and its impact on downstream tasks is well-documented in the CV literature (detailed discussion in Appendix \ref{appendix_1:related_work_general_cv}). However, this relationship for GFMs has received limited attention.

Previous GFMs used heuristic measures of diversity to decide where to sample pre-training data, with large differences from model to model. Presto developed a stratified random sampling strategy across hemispheres, ecological biomes, and land cover types \cite{tseng2023lightweight}. MMEarth sampled pre-training data uniformly across biomes and from five years \cite{nedungadi2024mmearth}. SatCLIP used a global uniform at random sampling \cite{klemmer2023satclip}. SeCo, SSL4EO, and CROMA used pre-training data within a range of 50km around 10,000 largest cities globally, assuming data will be more diverse around cities \cite{manas2021seasonal, wang2023ssl4eo, fuller2024croma}. Prithvi used pre-training data only from the contiguous United States \cite{jakubik2310foundation}. SatMAE, ScaleMAE, and SpectralGPT pre-trained on existing datasets fMoW \cite{christie2018functional}, which is biased towards the Global North, and BigEarthNet \cite{sumbul2019bigearthnet}, which includes only Europe \cite{cong2022satmae, reed2023scale, hong2024spectralgpt}.


Previous work chose different pre-training data sampling techniques due to differing assumptions about which strategy yields a diverse data distribution.
The impact of different sampling methods on downstream task performance remains unclear, as there has been no systematic comparison across different pre-training sampling techniques. In this research, we aim to fill this gap by providing empirical insights and guidance for future GFM development.

\section{Experimental Setup}
\label{sec:method}

\subsection{Task Formulation}
\label{subsec:task_formulation}

Our goal is to investigate how different pre-training data compositions affect the performance of GFMs on downstream tasks. We performed experiments with two GFMs (explained in \textsection \ref{subsec:geofm}), each pre-trained on five different data compositions created by various sampling techniques from globally available data, and compared with a baseline of no pre-training (details in \textsection \ref{subsec:pretraining_data_composition}).

For finetuning, we created subsets of downstream data samples based on their continent of origin. We split each continent subset into training, validation, and testing sets. We then finetuned each pre-trained model on task-specific training data from each continent (Africa, Asia, Europe, North America, Oceania, and South America) to investigate each model's performance in few-shot settings. This continent-wise approach allowed us to assess the model's effectiveness with region-specific tasks, which capture unique geographic and continental variations. We finetuned and evaluated each GFM on a specific downstream task explained in \textsection \ref{subsec:geofm}.

Since collecting labeled data for geospatial applications can be expensive \cite{rolf2024mission}, there is a need for models that can achieve good performance in a few-shot setting. We finetuned all downstream tasks on a few-shot learning setting by randomly selecting a total of $n=100$ samples from each continent's training set \cite{schick2020s, pecher2024comparing}.

To improve the reliability of the results for few-shot finetuning, we repeated each finetuning experiment 50 times with a different random seed for sampling training data. We reported the average and standard deviation across these iterations for each continent. We finetuned and evaluated each task using both parametric (Multi-Layer Perceptron (MLP) or logistic regression) and non-parametric (Random Forest and K-Nearest Neighbors (KNN)) models. Details of the experimental configuration are provided in Appendix \ref{appendix_3:experiment_details}.

\subsection{Geospatial Foundation Models and Downstream Tasks}
\label{subsec:geofm}

\paragraph{Presto} Presto is a lightweight pixel timeseries foundation model that aims to capture temporal patterns and spatial features efficiently \cite{tseng2023lightweight}. It is one of the few GFMs that uses time series satellite data to capture seasonal patterns. Presto uses various spectral bands from different satellite sensors, topographic data, location coordinates, etc.

We evaluated Presto using the CropHarvest agriculture downstream task \cite{tseng2021cropharvest}. We chose CropHarvest because it was used for evaluation in the original Presto paper and has global coverage. CropHarvest is a binary crop vs. ~non-crop classification task. We created a custom task from CropHarvest instead of using the pre-defined benchmark tasks to provide continent-wise subsets for finetuning (explained in the Appendix \ref{appendix_3:experiment_details}).

\paragraph{SatCLIP} SatCLIP is a location encoder-based foundation model \cite{klemmer2023satclip}, with a strong capability to learn implicit representations of locations. As a model optimized for extracting location-based features from satellite imagery, it effectively captures geographic and environmental characteristics. Being a location encoder, the input of each task consists of raw latitude and longitude coordinates.

We evaluated SatCLIP with the EcoRegions downstream task \cite{ecoregions} which is a multi-class classification of 14 classes. For this task, the model classifies ecological zones into predefined biome categories. This task assesses the model's understanding of geographical and ecological patterns, which include climate, soil type, vegetation, and biodiversity. We chose the EcoRegions task because it was used for evaluation in the original SatCLIP paper and has global coverage.

\subsection{Pre-training Data Compositions}
\label{subsec:pretraining_data_composition}

To vary the geographic distribution of pre-training data, we implemented several pre-training data sampling strategies, ranging from uniform random sampling to targeted regional geographies. Each technique was selected to capture different geographic data variability that could impact GFM performance. We experimented with five different data compositions, and zero pre-training as a baseline, detailed below:

\input{tables/result_mean_2_std_2}

\begin{itemize}
    \item Zero pre-training (Zero): To establish a baseline for assessing the impact of pre-training data, we employed zero pre-training in which the GFM is finetuned from random weights. 
    This measures the model's performance solely based on its architecture and finetuning data.
    \item Uniform at Random (UAR): To capture global representation, we uniformly and randomly sampled data with equal likelihood from the Earth's landmass. This sampling approach was used in \citet{klemmer2023satclip} to minimize bias toward any specific region or environment.
    \item Stratified Continent: We sampled pre-training data stratified by continents to achieve balanced representation across continental boundaries \cite{WorldContinents}. Within each stratum, we used uniform at random sampling to select points. We excluded Antarctica due to its limited land use and diversity.
    \item Stratified Biome: To capture environmental biodiversity and landcover, we uniformly sampled within each biome stratum, drawing an equal number of samples from the ecological zones created by \citet{ecoregions}.
    \item Natural Forest: To evaluate the model's capability in a strongly biased data context, we sampled all data exclusively from global intact forest cover \cite{intactForest}. This simulates cases where researchers may develop domain-specific foundation models, e.g., for forestry-related tasks.
    \item World Cities: In this population-centric sampling strategy, we sampled randomly within a 50 km radius around the world's 10,000 most populated cities \cite{SimpleMapsWorldCities}.
    This emulates the approach used by \citet{manas2021seasonal, wang2023ssl4eo,fuller2024croma} which sampled using a Gaussian distribution around the same cities, assuming land cover/land use is most diverse near cities.
\end{itemize}

We created pre-training datasets for SatCLIP using the Microsoft Planetary Computer data catalog \cite{microsoft_open_source_2022_7261897}. For Presto, we drew samples from the global pool of $\sim 22$ million samples provided by \cite{tseng2023lightweight}. We provided further details on the composition of each pre-training dataset, including the number of samples and visualizations of their geographic distributions, in Appendix \ref{appendix_2:pretraining_data_compositions}.

\section{Results and Analysis}
\label{sec:results}

Table \ref{tab:results} shows the F1 score for the CropHarvest and EcoRegions tasks using Random Forest (results of other finetuning models are included in Appendix \ref{appendix_4:extra_results}). We summarize our results in response to specific research questions below.

\begin{figure}[!ht]
    \centering
    \subfloat{\includegraphics[width=0.84\linewidth]{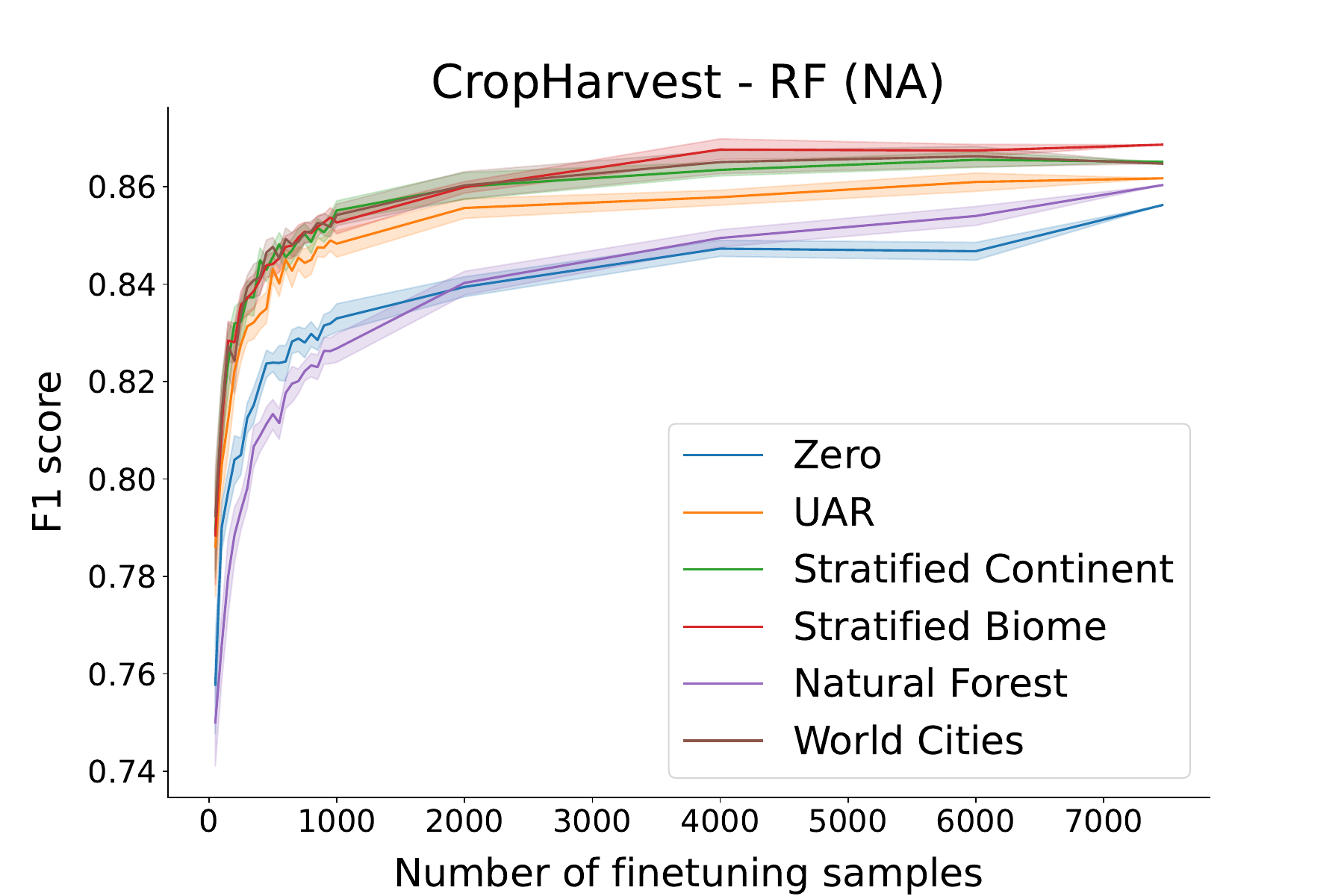}}\\
    \subfloat{\includegraphics[width=0.84\linewidth]{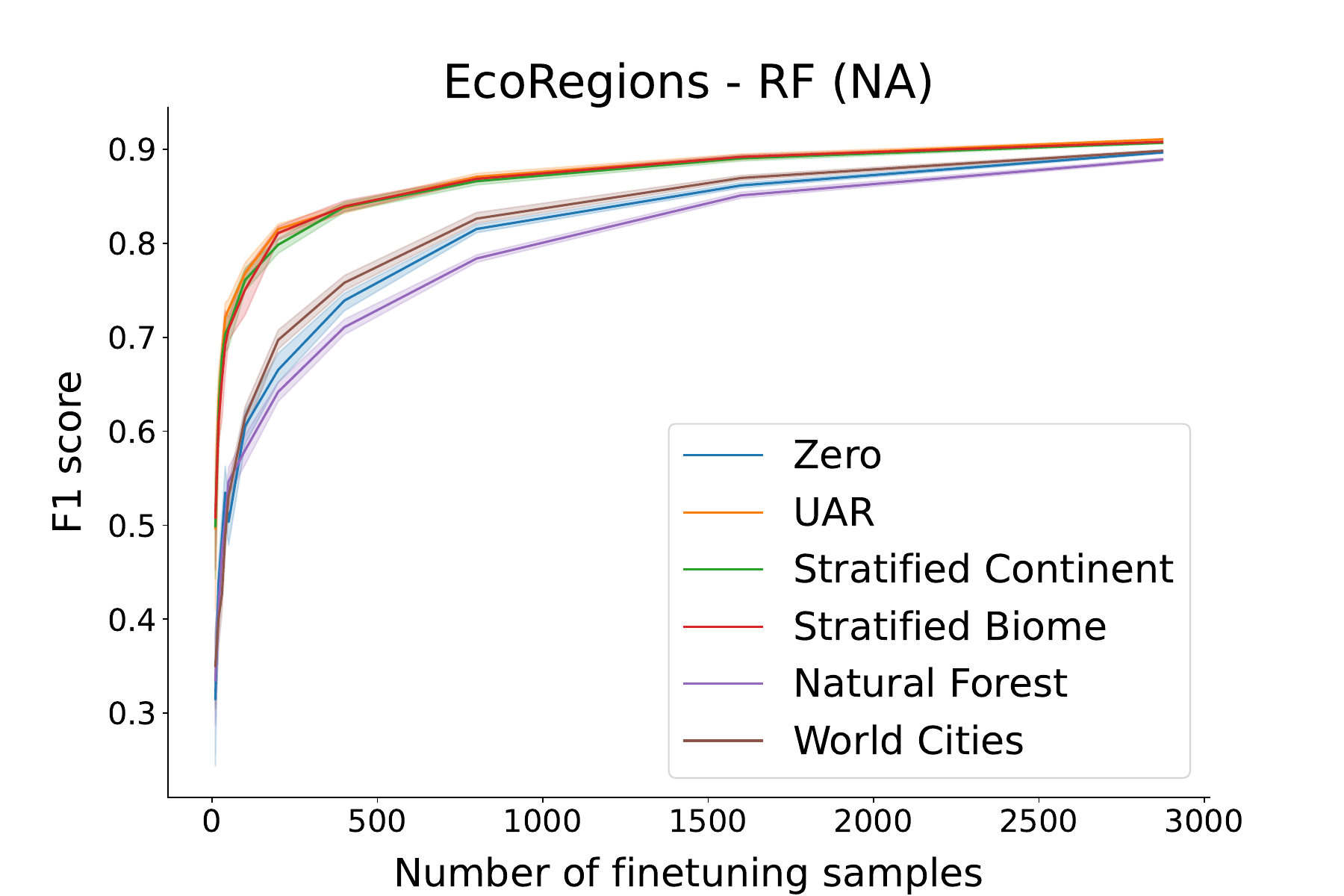}}
    \caption{Performance Comparison across different data compositions for North America (NA).}
    \label{fig:result_2}
    \vspace{-4mm}
\end{figure}

\textbf{How do different data compositions affect the performance of downstream tasks?}
Table \ref{tab:results} shows that models pre-trained with any data composition generally outperform models with zero pre-training. Overall, all balanced data sampling techniques (i.e., UAR, stratified continent/biome) outperform clustered techniques, i.e., Natural Forest for CropHarvest and both Natural Forest and World Cities for EcoRegions. Notably, performance is approximately consistent among balanced techniques, suggesting that \textit{among sampling strategies that ensure global representation, there is less difference in downstream task performance}. This finding can be further strengthened in future work by evaluating more balanced and clustered sampling techniques and additional downstream tasks. In CropHarvest, for South America and Oceania, Natural Forest performs on par with other data compositions, likely because Natural Forest covers a large portion of these continents.


\textbf{Do all data compositions show similar trends across different GFMs?}
In other words, can we expect that a specific data composition performs similarly across different models compared to other compositions? Our results indicate that the relative order of performance for different data compositions varies for each model. Table \ref{tab:results} shows that clustered data compositions, i.e., Natural Forest and World Cities, perform poorly in EcoRegions, however, World Cities shows an equal level of performance with other data compositions in CropHarvest. The reason behind this discrepancy could be the architectural design of the model. SatCLIP, which is a location-encoder model, struggles to generalize when pre-training data lacks global representation. However, Presto is a pixel timeseries model (which does not rely solely on location) and is likely more capable of generalizing on World Cities -- which is less clustered than Natural Forest. This hypothesis can be investigated in future work by performing experiments with additional GFMs based on location-encoding and pixel timeseries.

\textbf{Do performance differences between data compositions disappear when models are finetuned with an increasing amount of data?} Our findings indicate that pre-training data compositions impact how models perform with different amounts of finetuning data. Balanced data compositions outperform clustered data sampling in few-shot scenarios. However, when finetuning is conducted on an increasing amount of data (thousands of samples), the performance differences between various data compositions drop significantly. As shown in Figure \ref{fig:result_2} for North America, lines representing different data compositions converge as the number of finetuning samples increases (results of other continents are presented in the Appendix). Similar findings were observed across nine downstream tasks with ImageNet or LAION pre-training \citet{entezari2023role}.


\section{Conclusions}
\label{sec:conclusion}

As the development of GFMs advances, focusing on quality of pre-training data is essential, yet the impact of pre-training data distribution on downstream performance remains underexplored. To address this gap, our study is the first to investigate how pre-training data distribution can affect downstream task performance for GFMs. We conducted experiments on two GFMs by pre-training with different data compositions (including random, stratified, and clustered sampling) and evaluating in few-shot settings.

Experimental results indicate that balanced and global representative sampling techniques generally outperform clustered or region-specific compositions, highlighting the importance of globally diverse pre-training data for GFMs. Results also show that the effects of data sampling techniques can vary depending on model architecture. Moreover, balanced sampling techniques exhibit nearly equal performance, suggesting that the choice of specific sampling techniques is insignificant, as long as the composition ensures balanced samples across the globe.

Our results provide preliminary insights into how the geographic distribution of pre-training data can influence downstream task performance. However, further exploration can be done to deepen our understanding by adding more GFMs and data sampling strategies. Detailed limitations and directions for future work are provided in the Appendix \ref{appendix_5:future_work}.



\bibliography{aaai25}

\appendix

\newpage

\section{Pre-training Data Distribution Study in General CV}
\label{appendix_1:related_work_general_cv}

The effect of pre-training data distribution on performance has been widely explored in general CV research. Hammoud et. al. shows that increasing pre-training data diversity improves the performance of the self-supervised model, but only when the distribution closely aligns with the downstream task data \cite{hammoud2024pretraining}. For an ImageNet-scale dataset, reducing the pre-training data by half (from 1M to 500k images) only decreases downstream performance by about 1-2\% \cite{cole2022does}. Another study \cite{ramanujan2024connection} shows that pre-training with additional data enhances robustness, however, a significant gain can be achieved without large datasets. They show that using only 25K images from ImageNet or iNaturalist—6x smaller than the finetuning dataset—already provides noticeable improvements. Cole et. al. and Ramanujan et. al. establish that simply increasing data volume without increasing diversity does not always benefit. \citet{longpre2023pretrainer} examines how pre-training data choices affect language model performance, showing that data age, quality, toxicity filtering, and domain diversity significantly influence outcomes. Longpre et. al. also demonstrates that diverse data sources, like books and web content, improve robustness in language model development. Previous work also shows that different pre-training distributions lead to variations in transfer accuracy and these differences are more significant in few-shot finetuning \cite{entezari2023role}. Having a variety of objects and their characteristics in pre-training data significantly improves the generalization capability of the CLIP model \cite{abbasiclip}. Also, \citet{fang2022data} shows that CLIP's robustness largely depends on the choice of training data, with other factors having minimal impact.

\begin{figure*}[!ht]
    \centering
    \subfloat[Zero]{\includegraphics[width=0.48\linewidth]{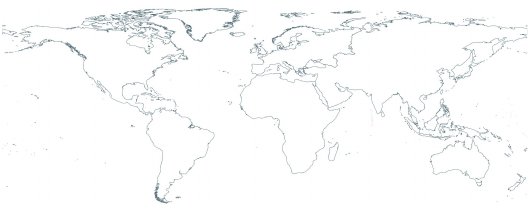}}
    \hspace{1em}
    \subfloat[Uniform at Random]{\includegraphics[width=0.48\linewidth]{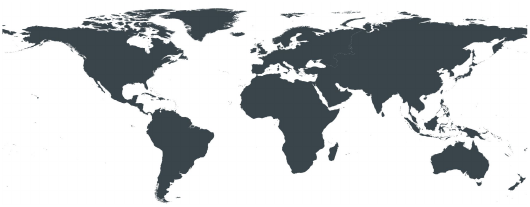}}\\[0.5em]
    \vspace{0.5em}
    \subfloat[Stratified Continent]{\includegraphics[width=0.48\linewidth]{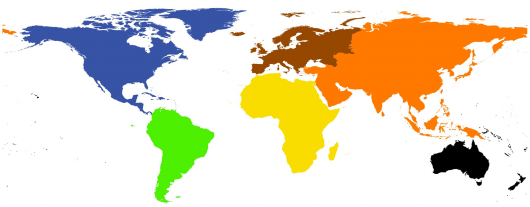}}
    \hspace{1em}
    \subfloat[Stratified Biome]{\includegraphics[width=0.48\linewidth]{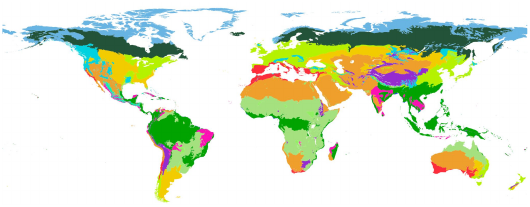}}\\[0.5em]
    \vspace{0.5em}
    \subfloat[Natural Forest]{\includegraphics[width=0.48\linewidth]{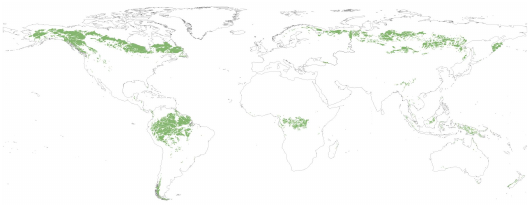}}
    \hspace{1em}
    \subfloat[World Cities]{\includegraphics[width=0.48\linewidth]{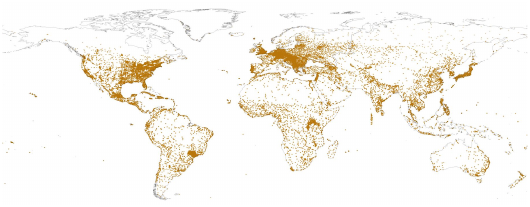}}
    \caption{Geographic distribution of pre-training data compositions}
    \label{fig:appendix_pretraining_data_composition}
\end{figure*}

\section{Details of Pre-training Data}
\label{appendix_2:pretraining_data_compositions}

As discussed in \textsection \ref{subsec:pretraining_data_composition}, to vary spatial distributions of pre-training datasets, we experimented with five data composition techniques each designed to capture varying levels of geographic coverage and regional characteristics as shown in Figure \ref{fig:appendix_pretraining_data_composition}. These compositions range from broad global diversity to targeted environmental and urban-focused regions, offering unique spatial distributions for analysis. To do a fair comparison across pre-training sampling strategies, we used an equal number of pre-training samples in all data compositions for a given GFM. Details of pre-training data distribution for each GFM are as follows:



\begin{itemize}
    \item \textbf{Presto}: For each composition, we resampled pre-training datasets from an existing pool of approximately $\sim 22$ million samples provided by the Presto dataset \cite{tseng2023lightweight}. Particularly, for each data composition, we sampled 6.5 million data points. The choice of 6.5 million samples was driven by the need to maintain equal samples across strata in stratified sampling techniques.
    \item \textbf{SatCLIP}: We created pre-training datasets by sampling 100,000 patches for each composition strategy, following the original SatCLIP data sampling configuration \cite{klemmer2023satclip}. We used Sentinel-2 data by excluding one band as specified in the SatCLIP implementation. To export data, we used Microsoft Planetary Computer catalog \cite{microsoft_open_source_2022_7261897}. The temporal range for sampling spanned from January 1st, 2021 to September 15th, 2024. The dataset for each sampling strategy was created from scratch to ensure the uniqueness of each composition.
\end{itemize}

\section{Experiment Configuration}
\label{appendix_3:experiment_details}

\textbf{Pre-training:} To pre-train Presto and SatCLIP, we used the same configuration and hyperparameters as specified in the original studies, utilizing the official repositories provided by \citet{tseng2023lightweight}\footnote{\url{https://github.com/nasaharvest/presto}} and \citet{klemmer2023satclip}\footnote{\url{https://github.com/microsoft/satclip}}, respectively.


\noindent \textbf{Downstream Tasks and Finetuning:} As described in \textsection \ref{subsec:geofm}, we finetune the CropHarvest and EcoRegions tasks continent-wise. To achieve this, we first categorize the data samples of each task into continent categories, Table \ref{tab:no_of_samples} provides the number of samples for all six continents and the total dataset size across both tasks. We then split the data from each continent into training, validation, and testing sets for finetuning and evaluation. We did not consider country-level finetuning and evaluation, as the country-level samples were insufficient, and the reduced sample size in the testing set prevented us from drawing reliable conclusions.

For CropHarvest, we modified the task from what was originally proposed in \cite{tseng2021cropharvest}. The original CropHarvest released an evaluation set focused on specific crop categories in three countries. Instead, we used the full CropHarvest dataset to create continent-specific subsets for finetuning. For each continent, we split the data into 70\% for training, 10\% for validation, and 20\% for testing. For EcoRegions, we used the dataset including its training, validation, and testing splits as it was employed in SatCLIP.

\begin{table*}
    \centering
    \caption{Total and continent-wise number of samples in downstream tasks}
    \label{tab:no_of_samples}
    \begin{tabular}{ccccccc|c}
        \toprule[1.5pt]
        & \textbf{AF} & \textbf{AS} & \textbf{EU} & \textbf{NA} & \textbf{OC} & \textbf{SA} & \textbf{Total} \\
        \midrule
        CropHarvest & 21693 & 14475 & 15784 & 10656 & 756 & 6812 & 70176\\
        EcoRegions & 4027 & 8333 & 2246 & 5747 & 1177 & 2344 & 23874\\
        \bottomrule[1.5pt]
    \end{tabular}
\end{table*}

As discussed in the \textsection \ref{sec:results}, we conducted finetuning and evaluation on both tasks using an increasing amount of data. For both downstream tasks, we finetuned the model on varying sample sizes, starting from $n=10$ up to the total available training samples in the continent. As the number of fineuning samples $n$ increases, we proportionally reduce the number of iterations $r$ used to compute the final average performance metric, i.e., $n \propto \frac{1}{r}$. Importantly, in all finetuning and evaluation, only the training set is changing, validation and test set remain fixed across all experimental configurations.

\section{Additional Results}
\label{appendix_4:extra_results}

As mentioned in \textsection \ref{subsec:task_formulation}, we evaluated each task using both parametric models (Multi-Layer Perceptron (MLP) and logistic regression) and non-parametric models (Random Forest (RF) and K-Nearest Neighbors (KNN)).

\begin{figure*}[!hb]
    \centering
    \subfloat{\includegraphics[width=0.32\linewidth]{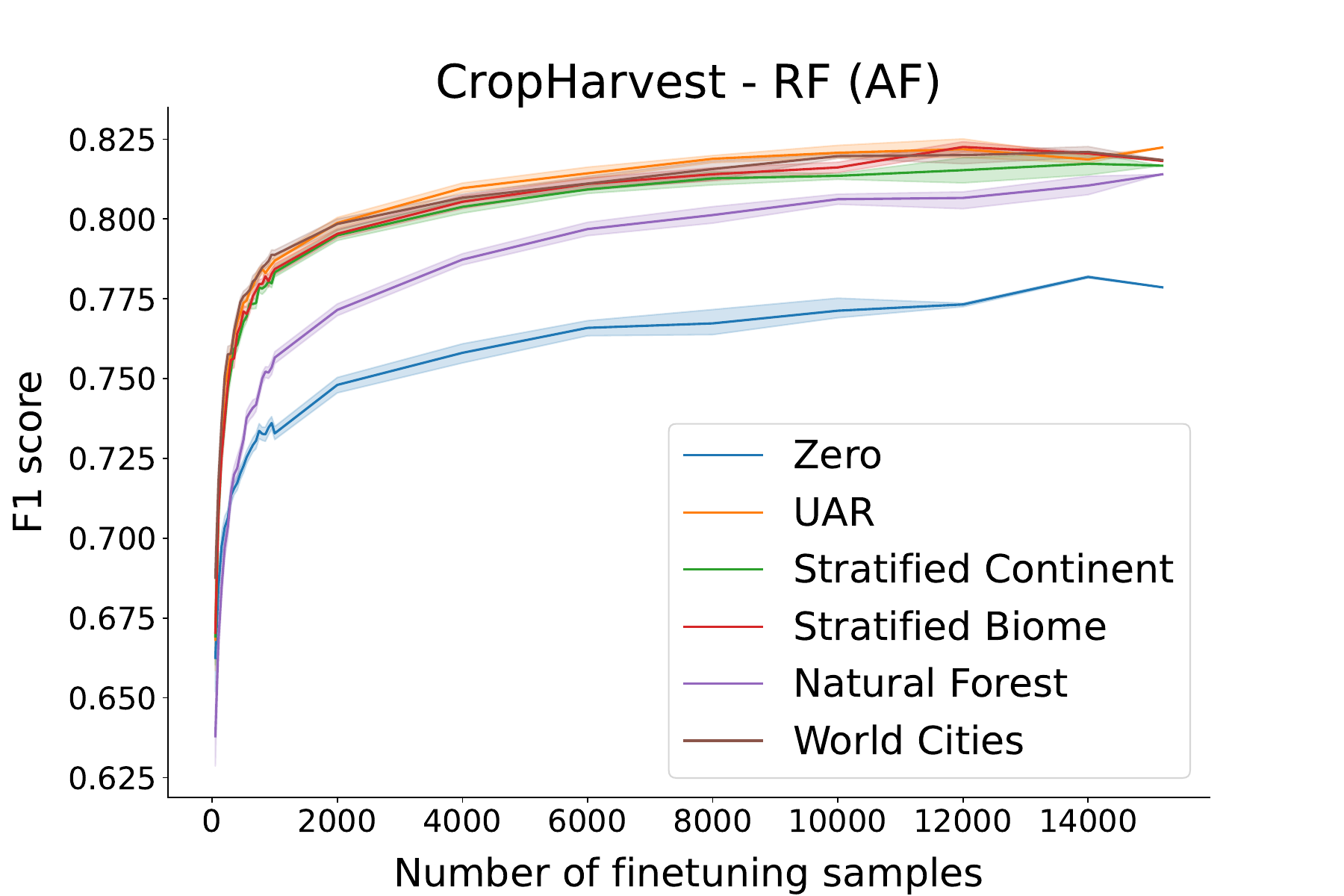}}
    \subfloat{\includegraphics[width=0.32\linewidth]{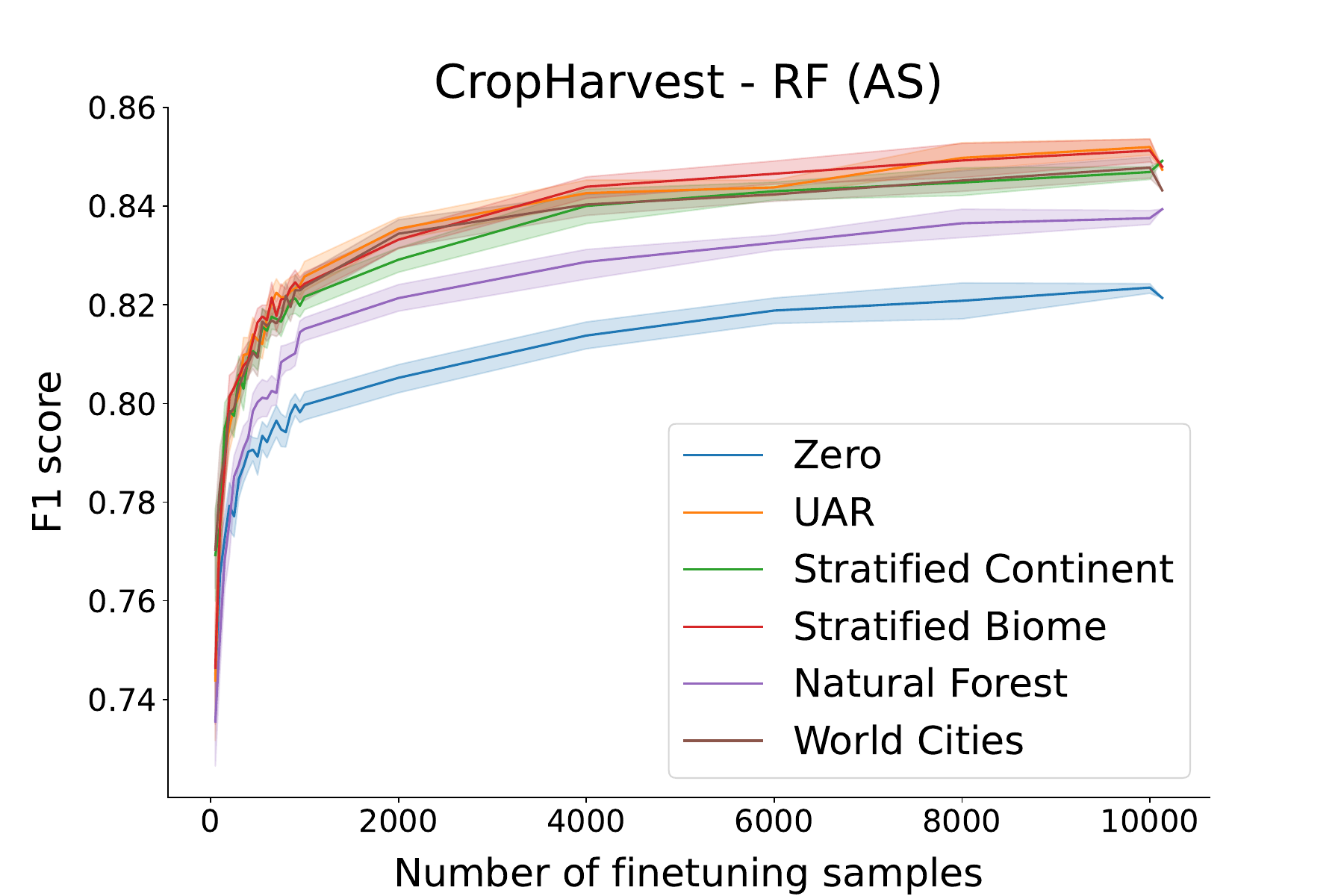}}
    \subfloat{\includegraphics[width=0.32\linewidth]{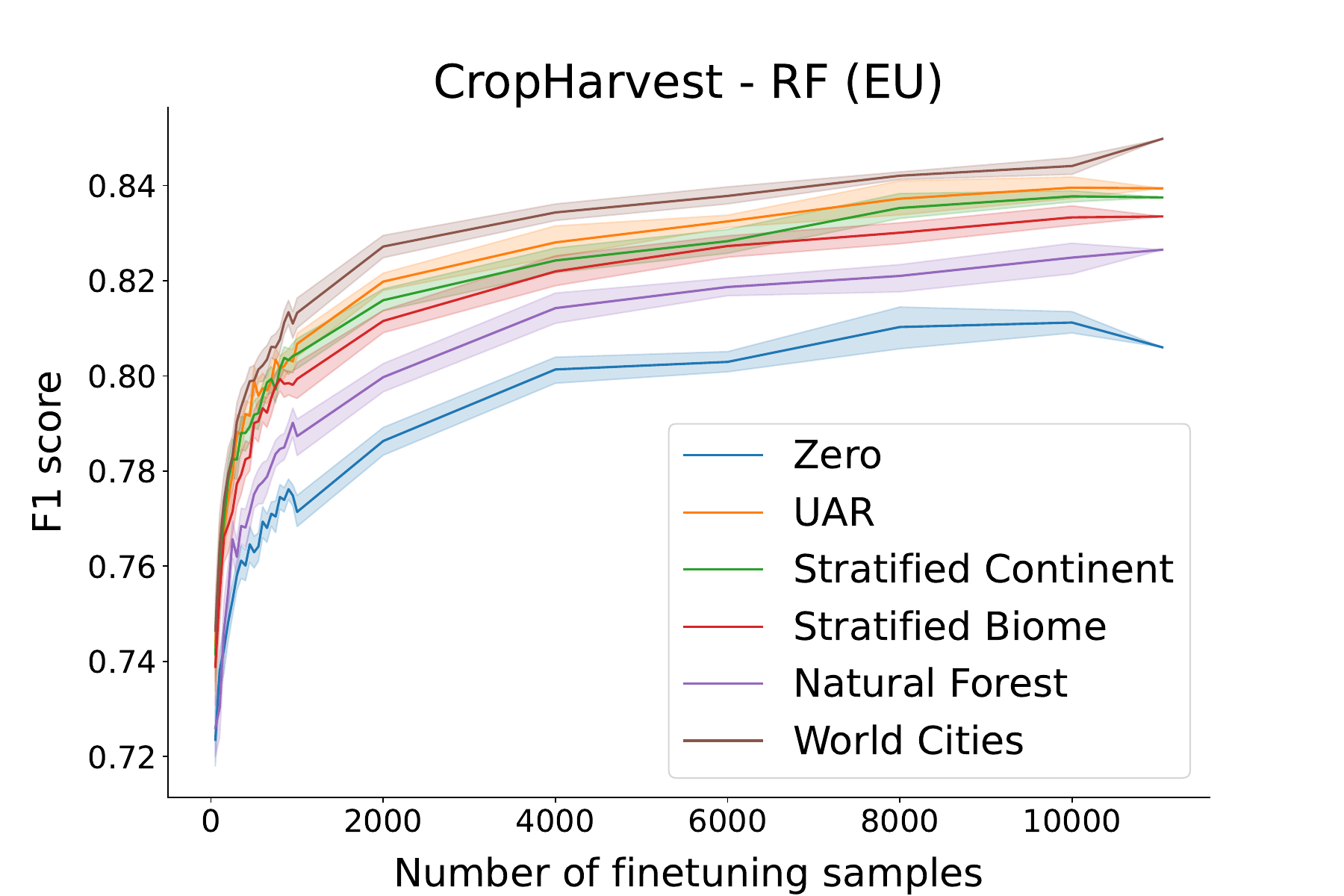}}\\
    \subfloat{\includegraphics[width=0.32\linewidth]{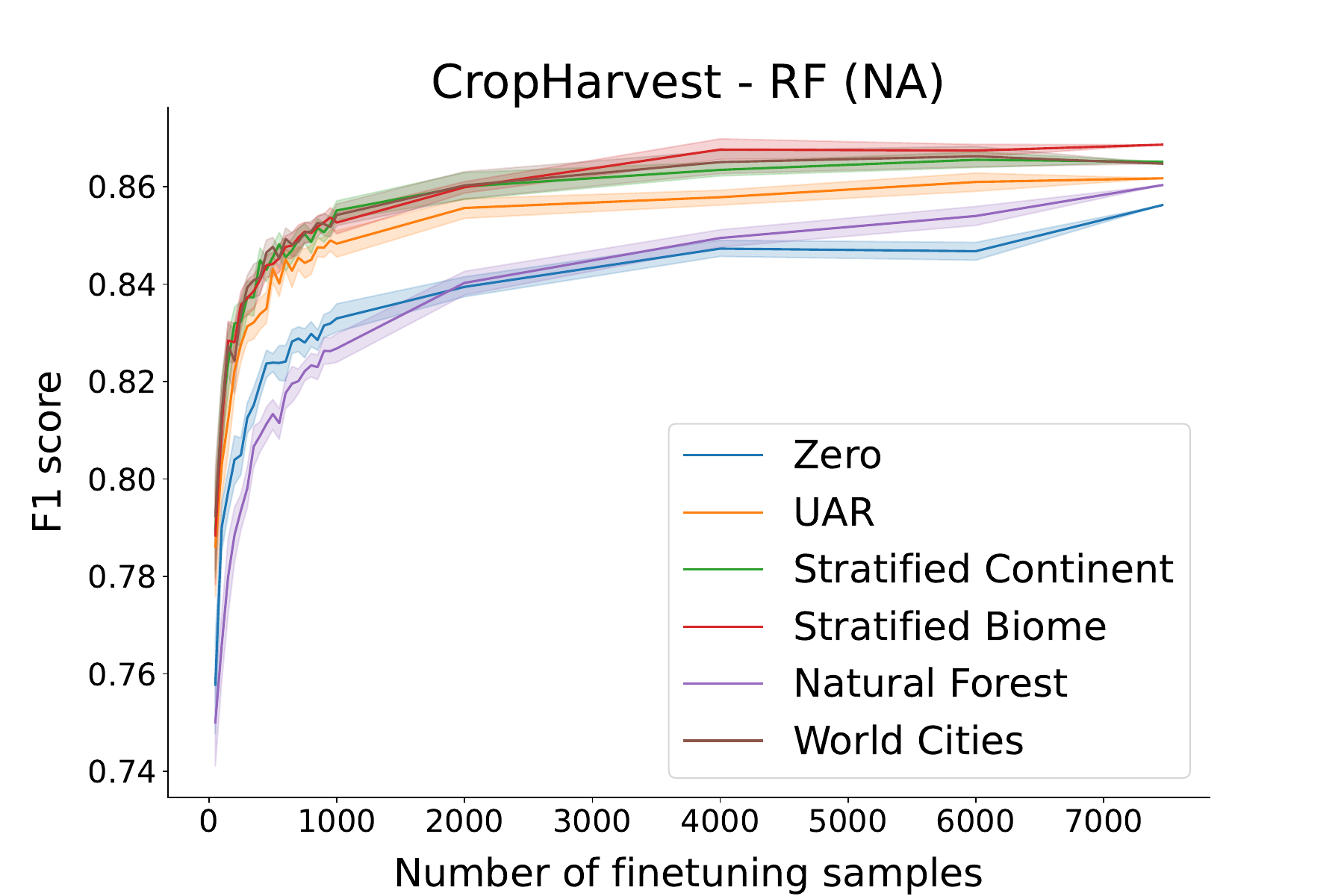}}
    \subfloat{\includegraphics[width=0.32\linewidth]{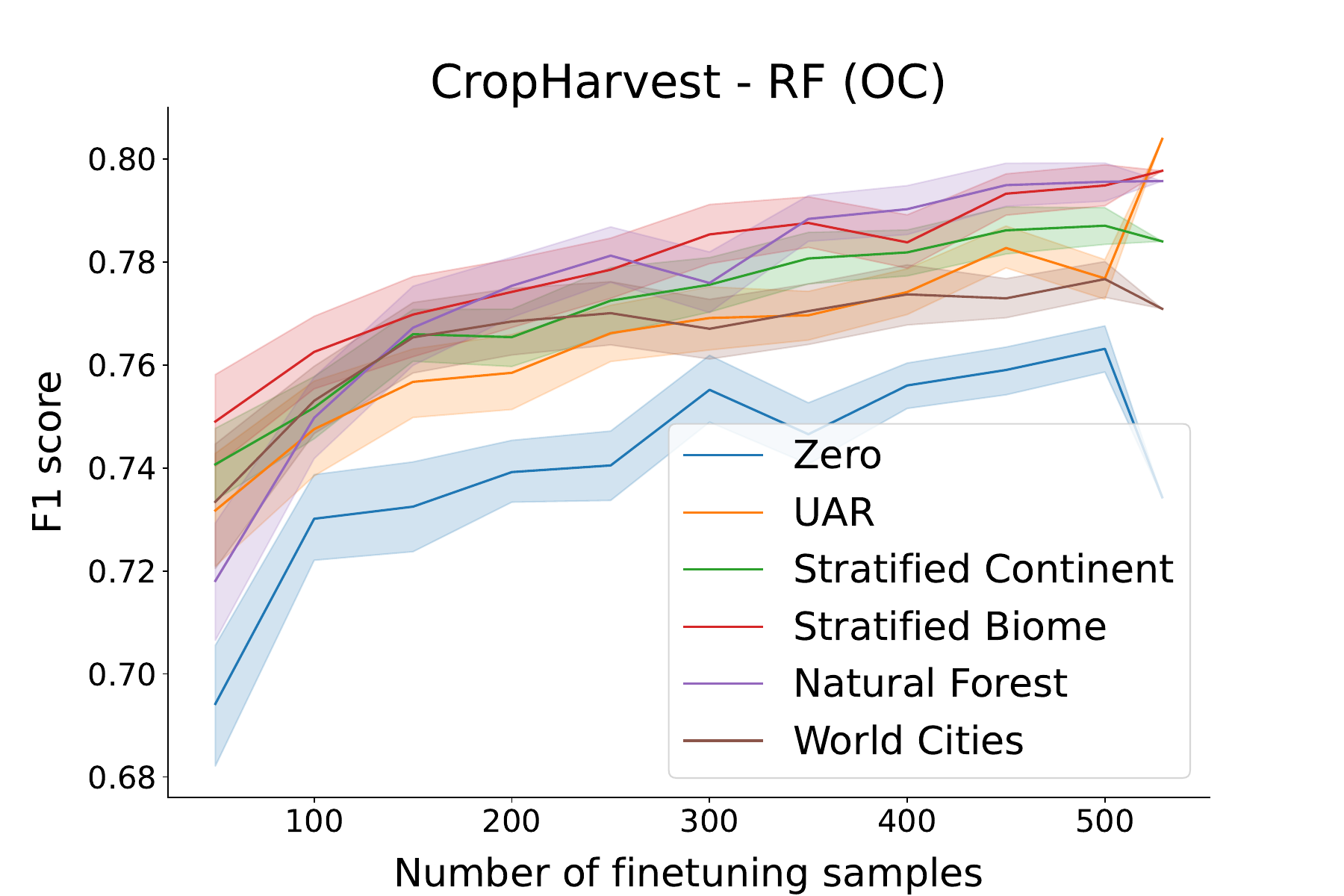}}
    \subfloat{\includegraphics[width=0.32\linewidth]{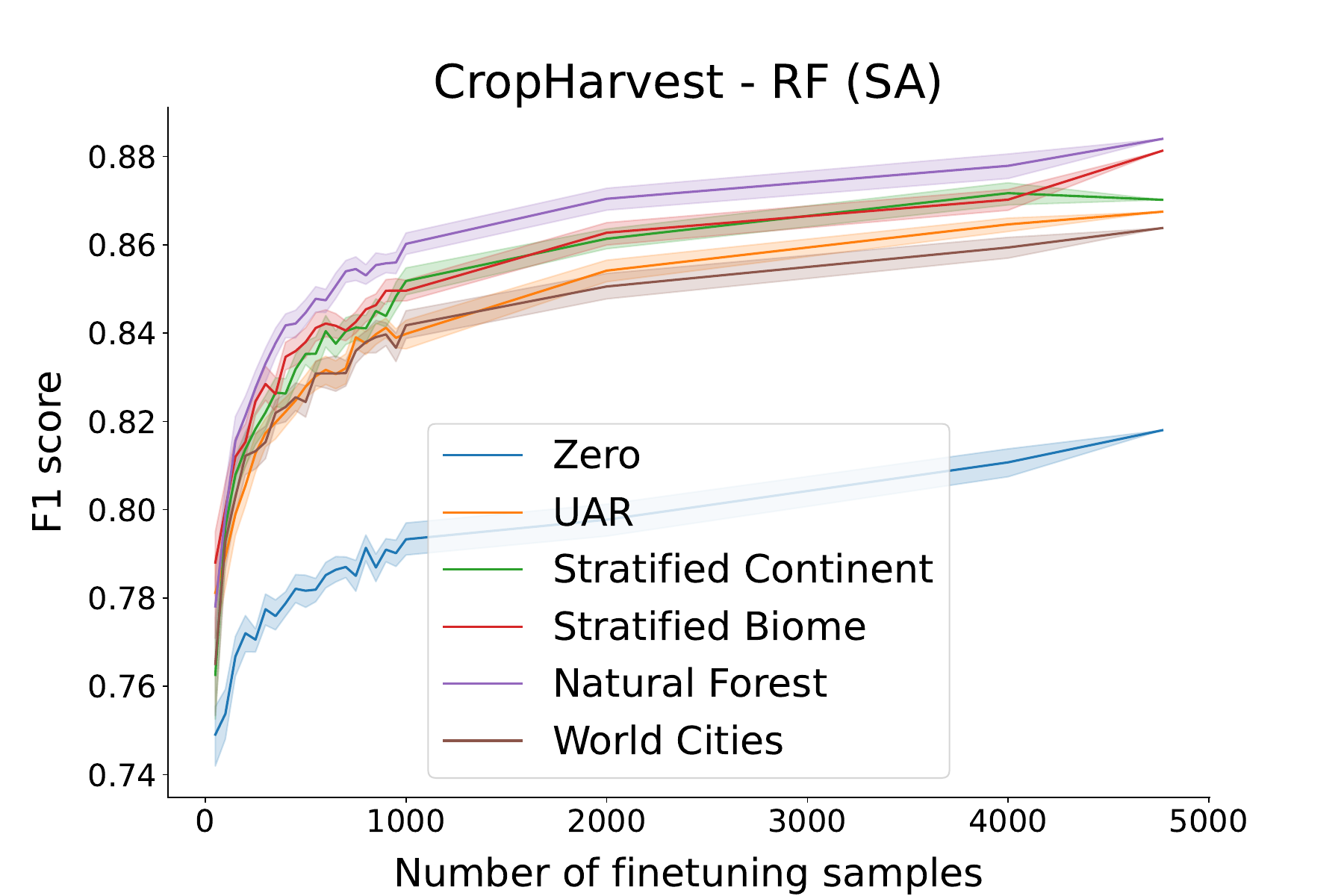}}
    \caption{Continent-wise results for CropHarvest task with Random Forest}
    \label{fig:result_cropharvest_rf}
\end{figure*}

\begin{figure*}[t]
    \centering
    \subfloat{\includegraphics[width=0.32\linewidth]{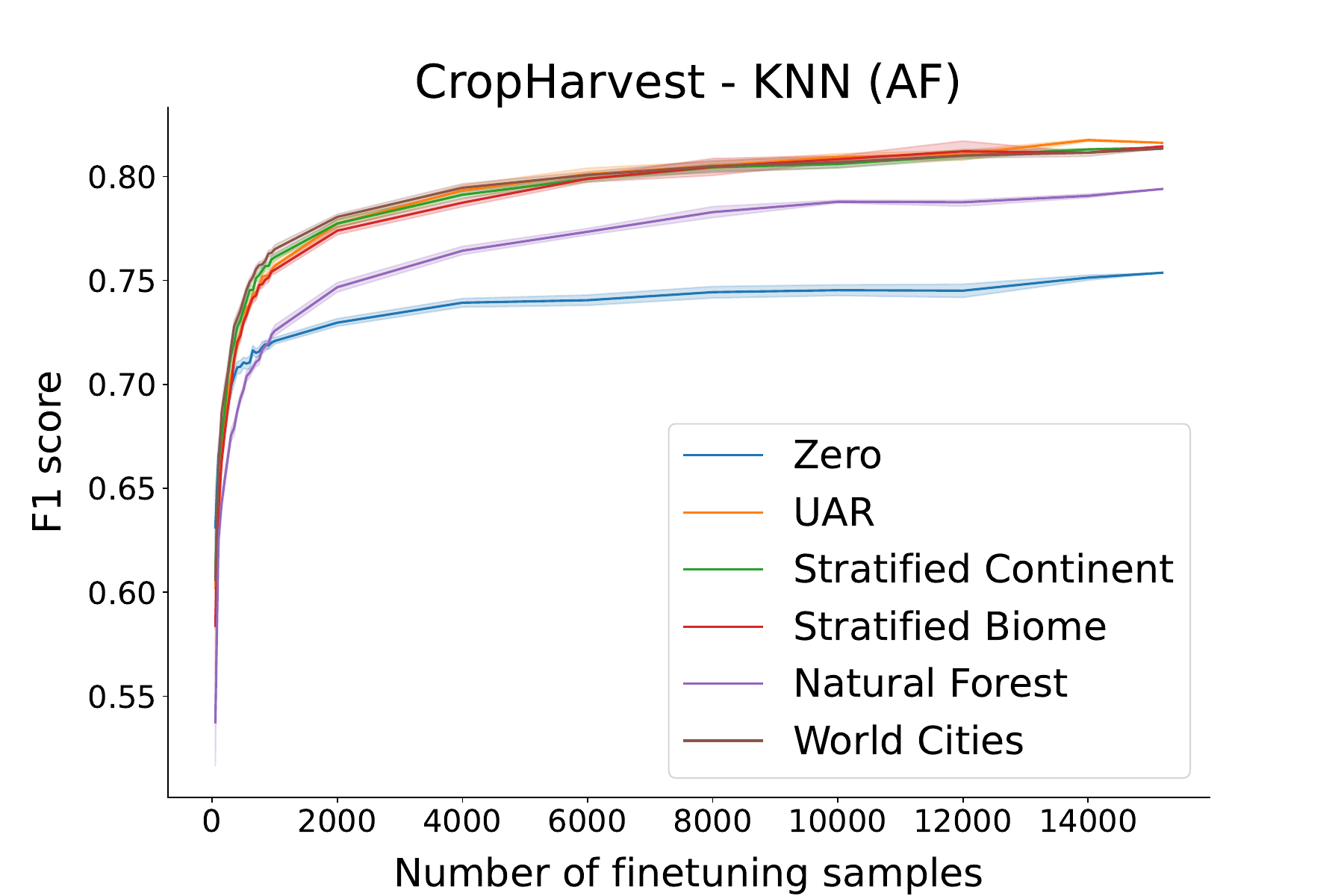}}
    \subfloat{\includegraphics[width=0.32\linewidth]{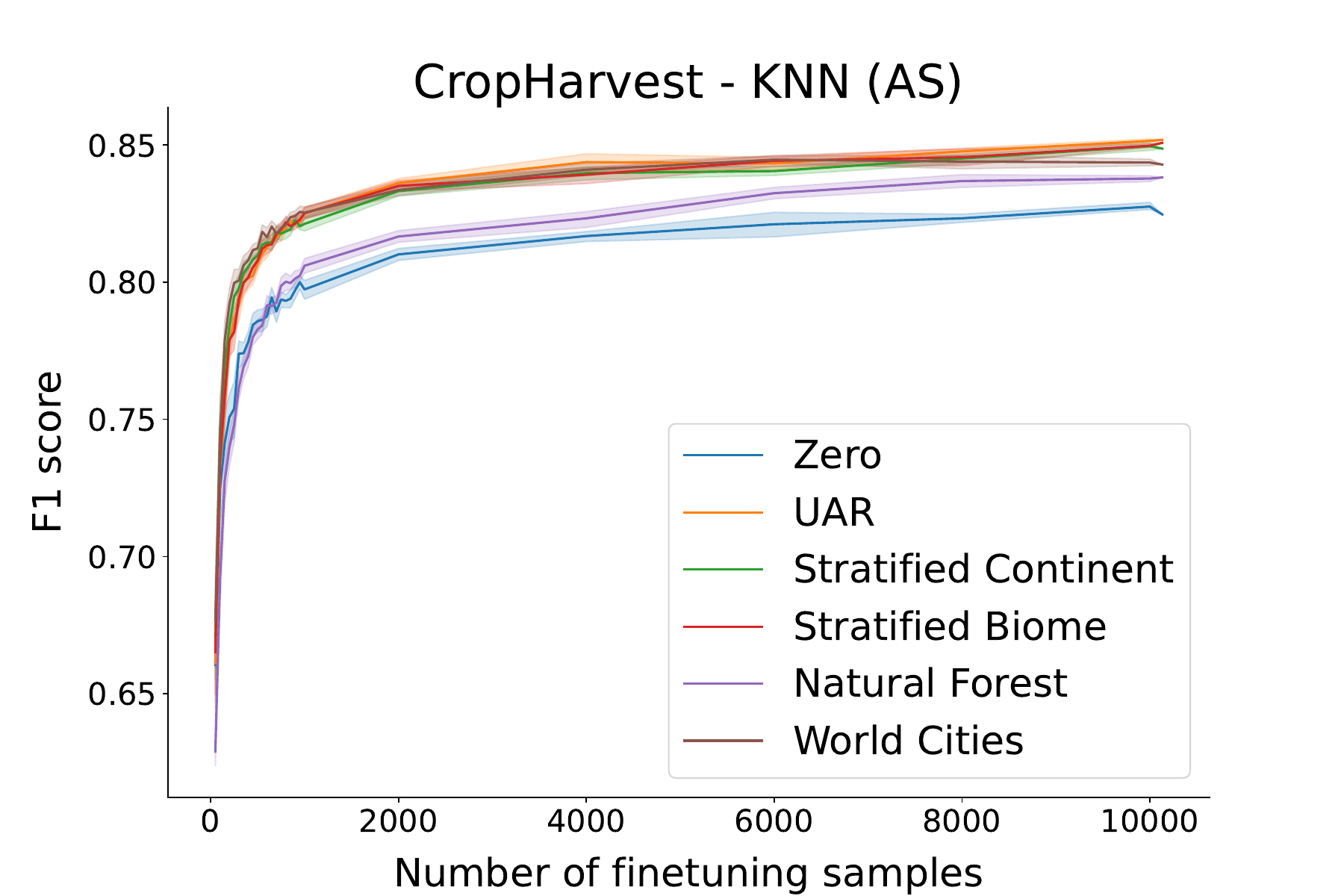}}
    \subfloat{\includegraphics[width=0.32\linewidth]{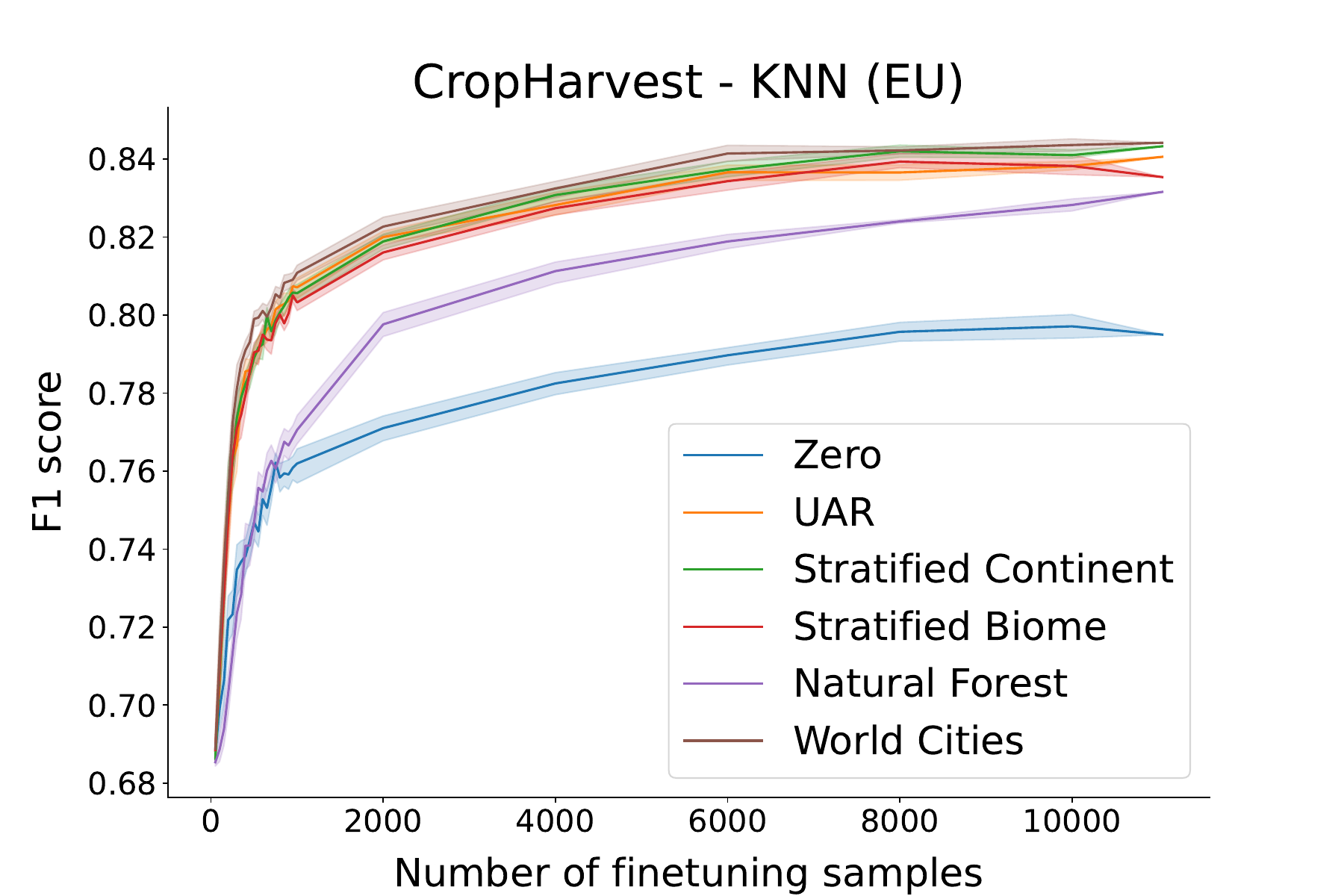}}\\
    \subfloat{\includegraphics[width=0.32\linewidth]{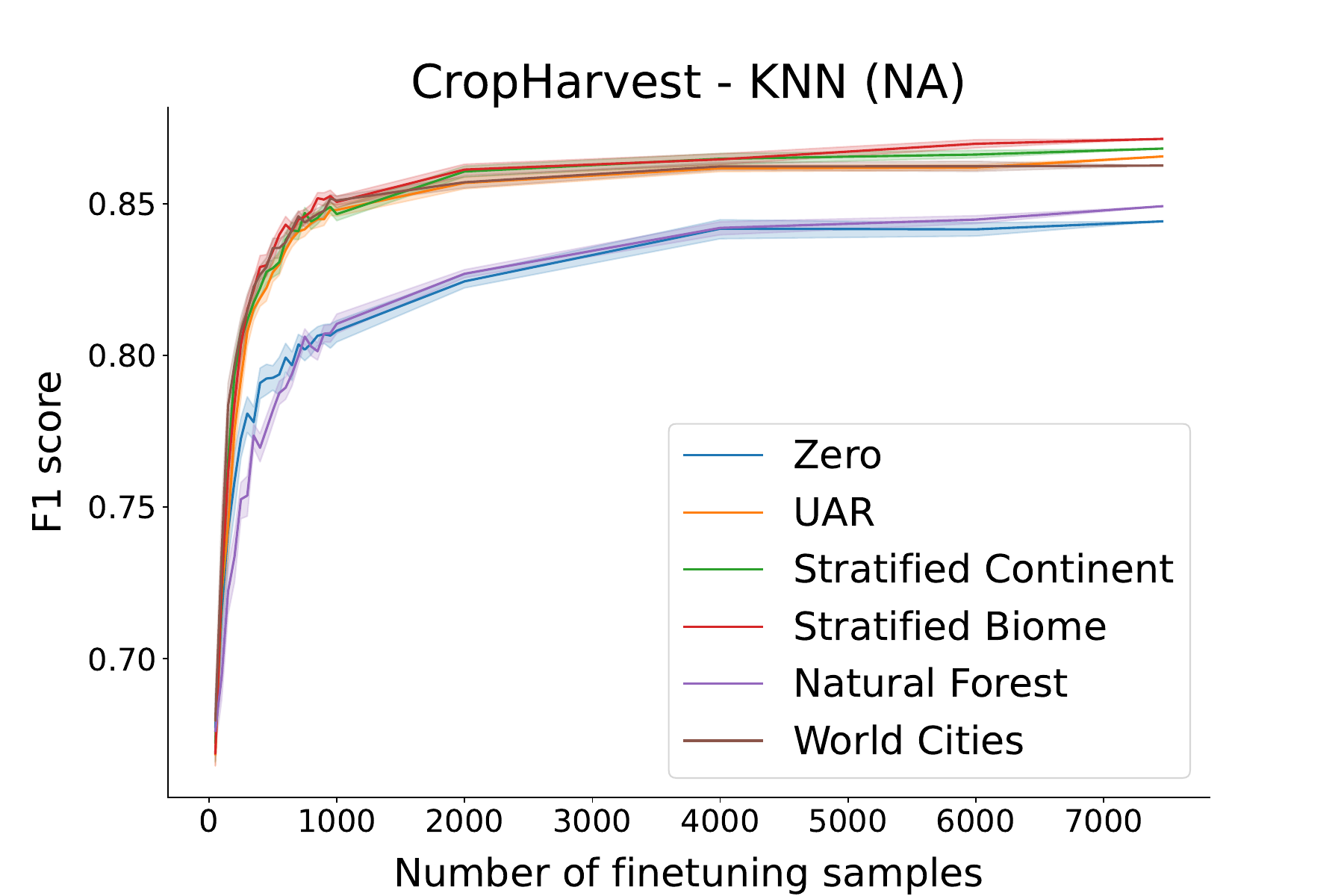}}
    \subfloat{\includegraphics[width=0.32\linewidth]{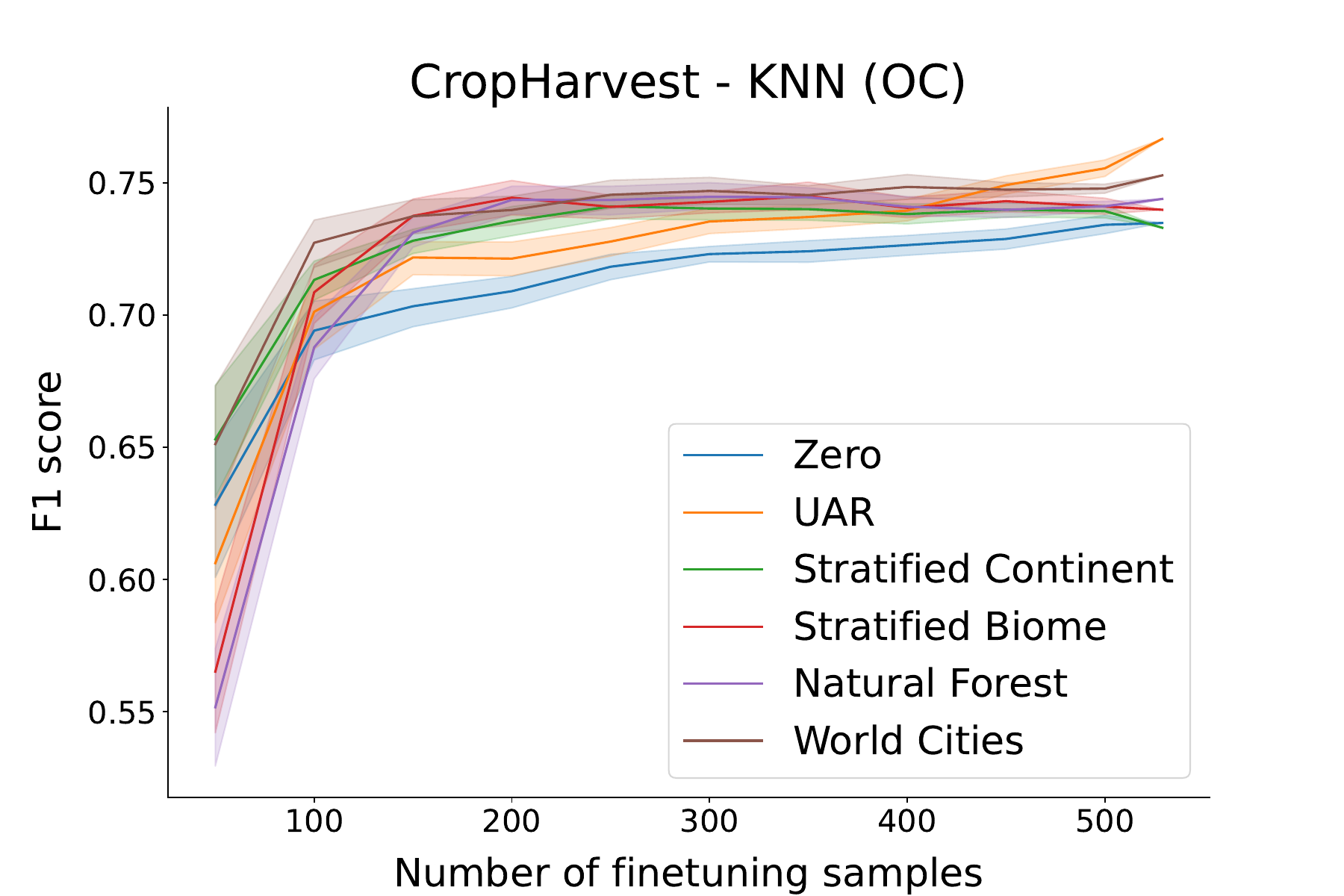}}
    \subfloat{\includegraphics[width=0.32\linewidth]{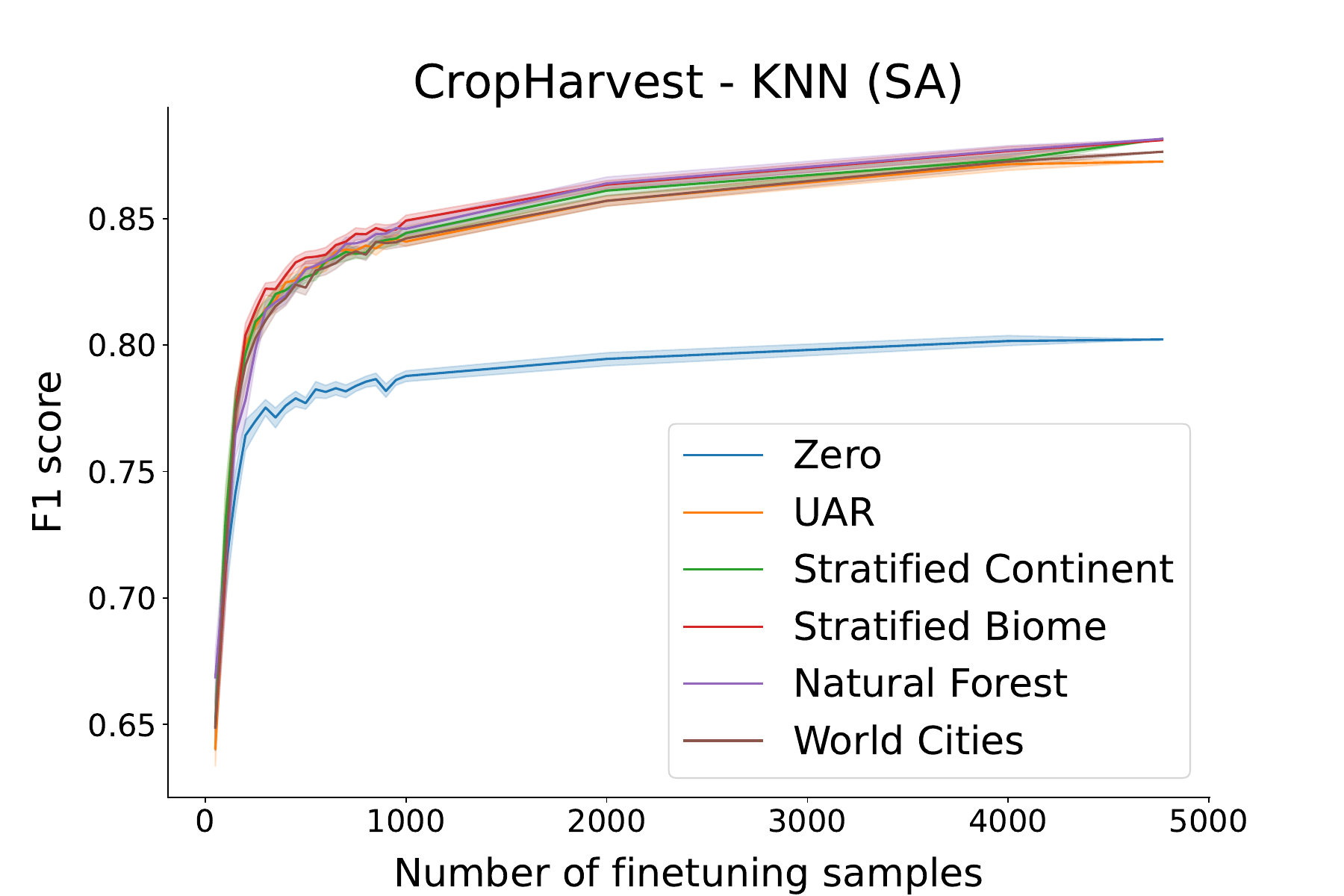}}
    \caption{Continent-wise results for CropHarvest task with KNN}
    \label{fig:result_cropharvest_knn}
\end{figure*}

\begin{figure*}[t]
    \centering
    \subfloat{\includegraphics[width=0.32\linewidth]{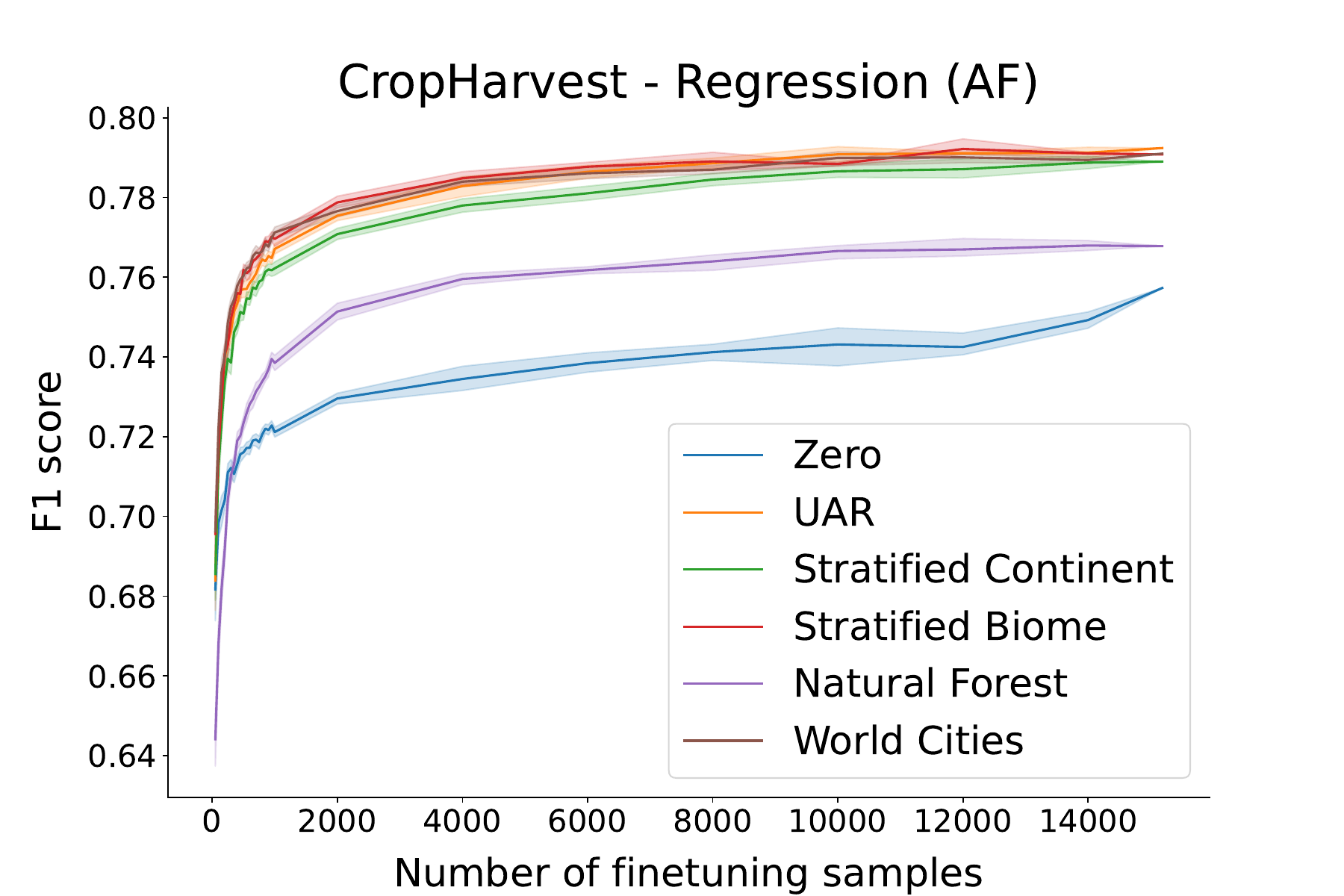}}
    \subfloat{\includegraphics[width=0.32\linewidth]{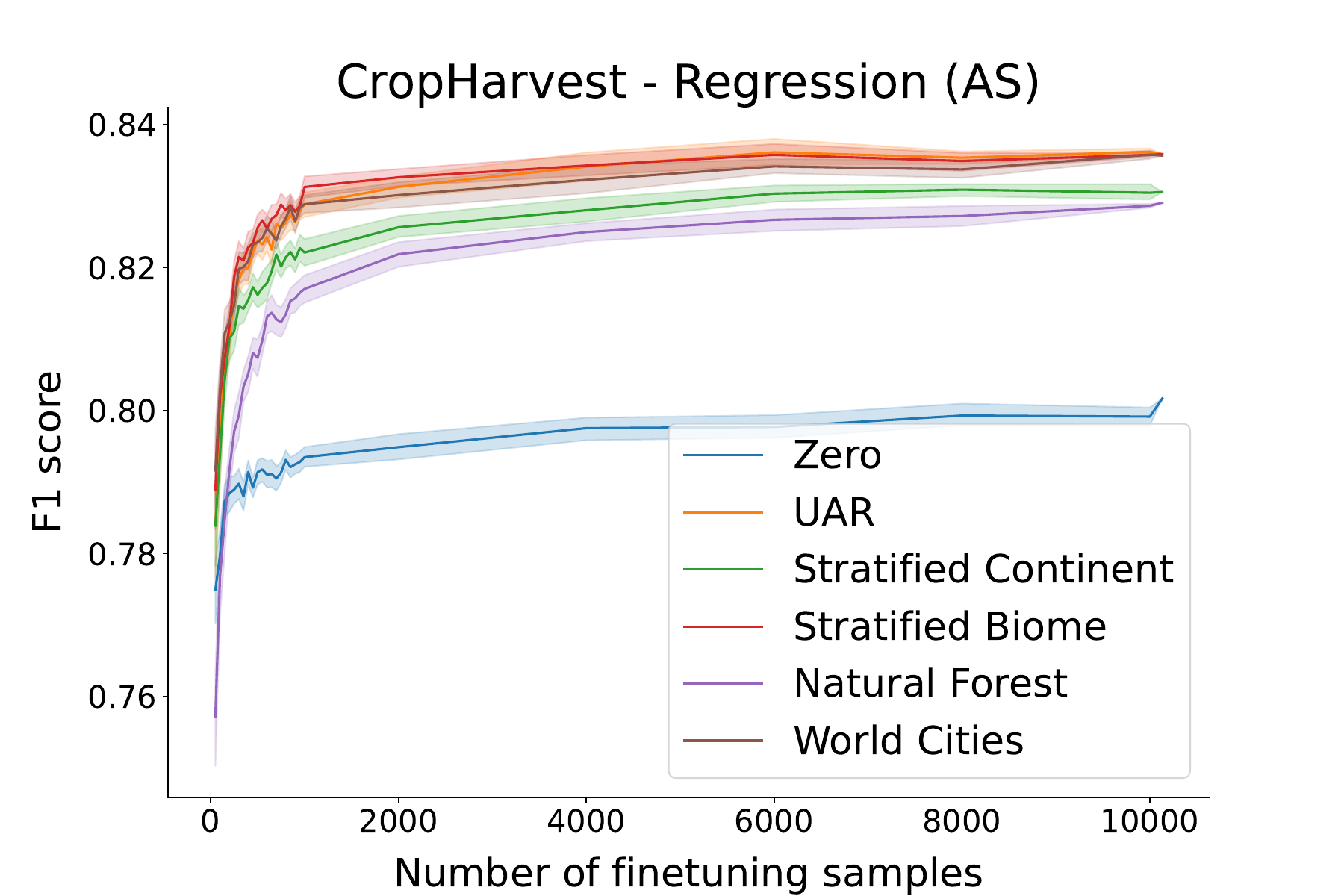}}
    \subfloat{\includegraphics[width=0.32\linewidth]{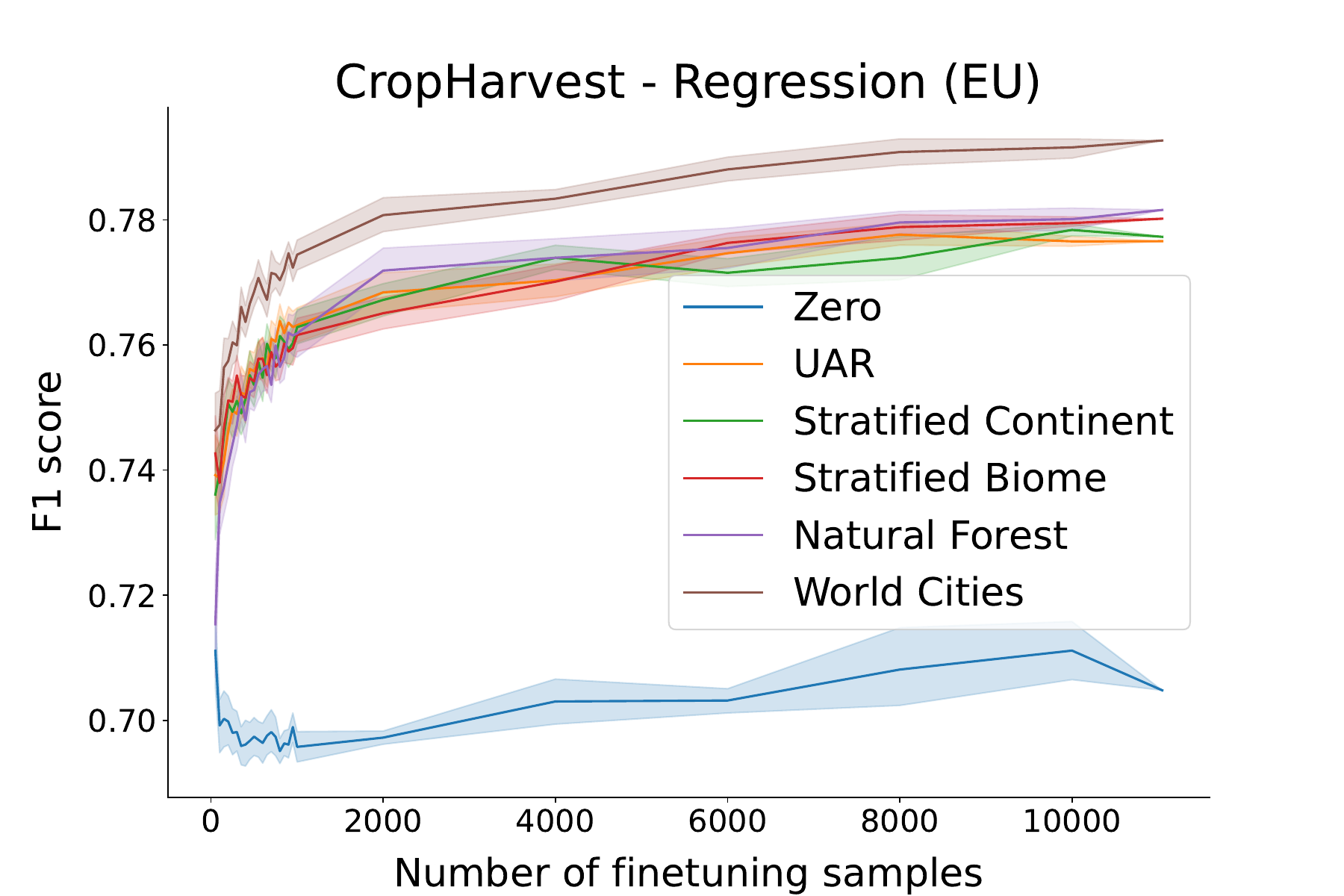}}\\
    \subfloat{\includegraphics[width=0.32\linewidth]{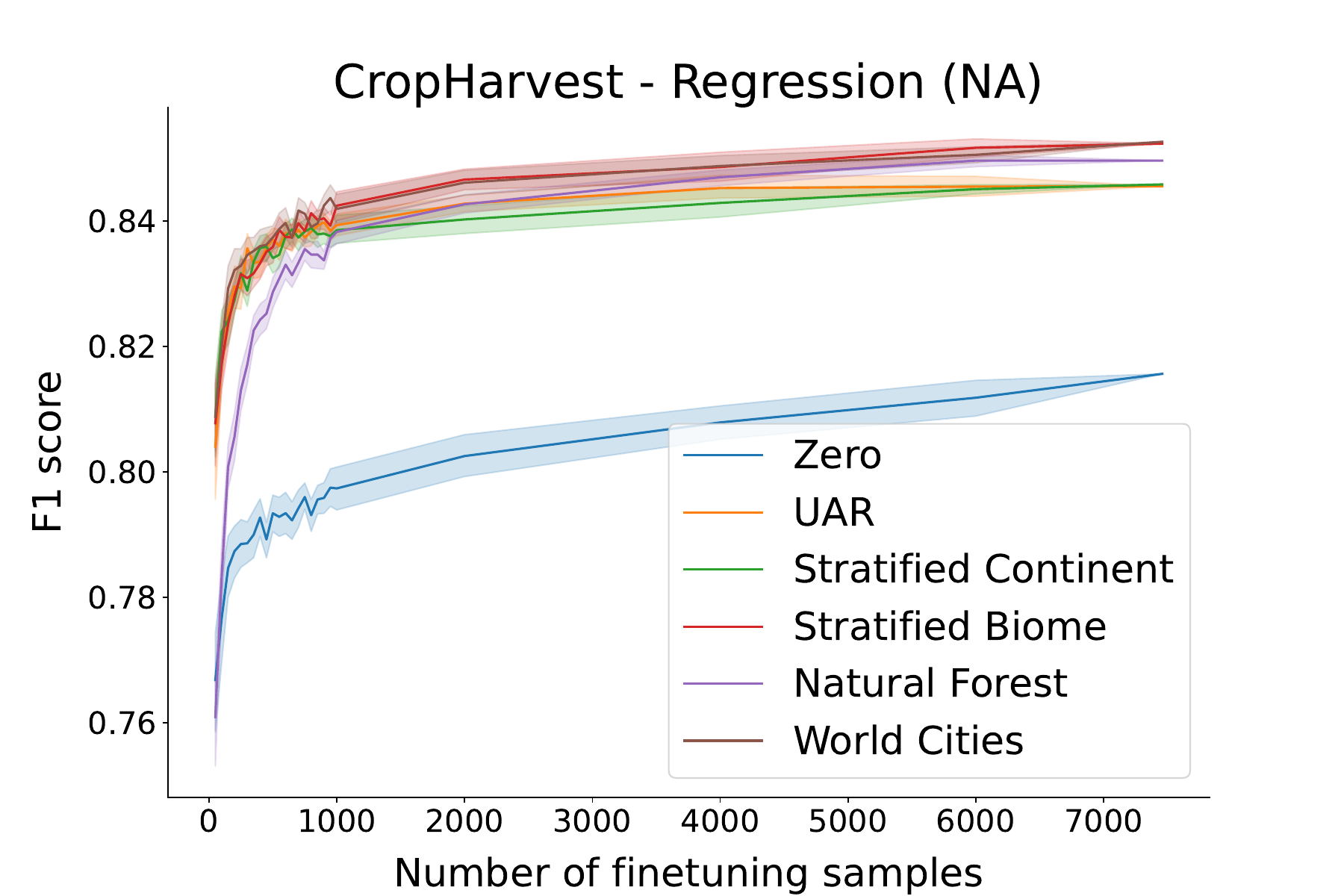}}
    \subfloat{\includegraphics[width=0.32\linewidth]{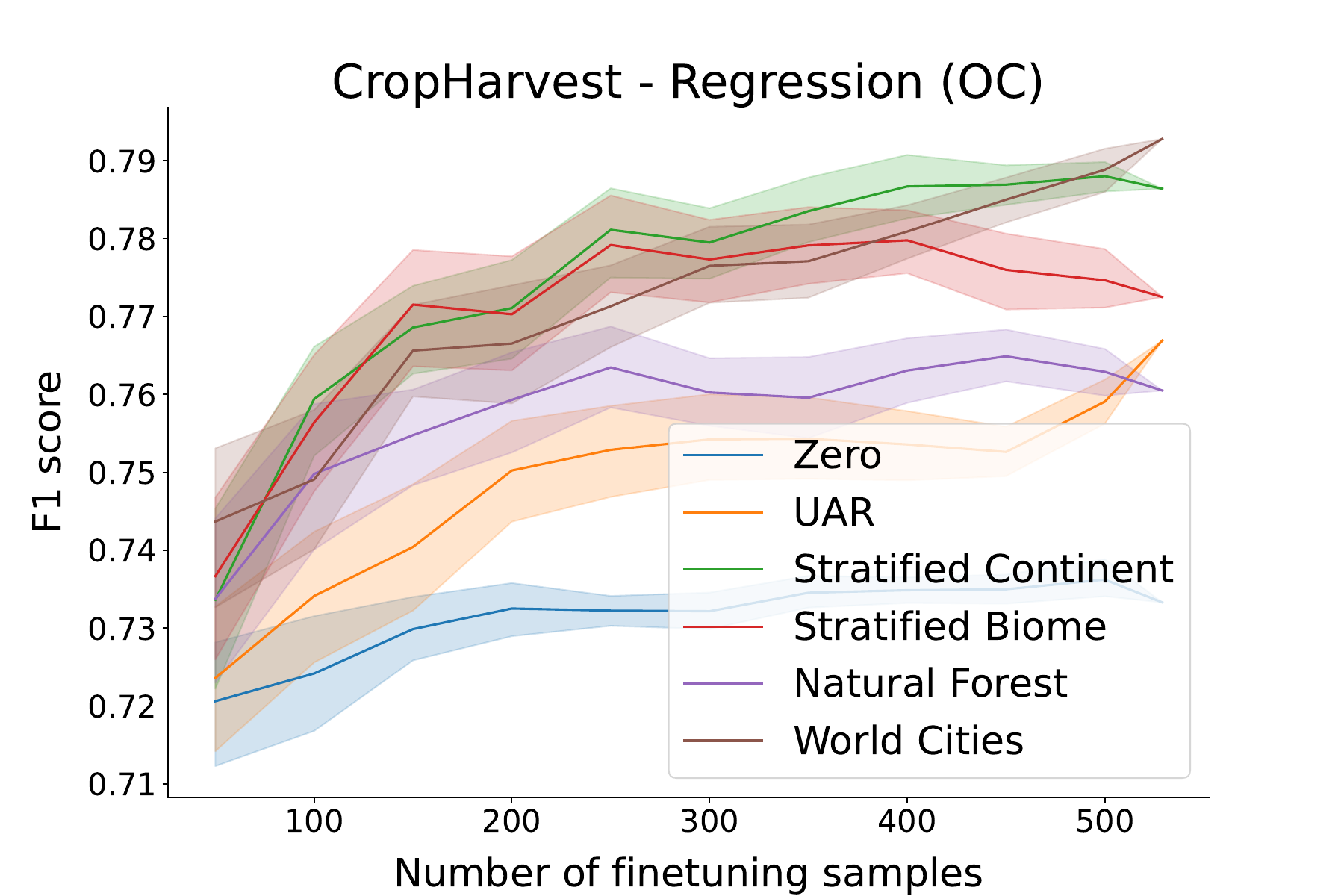}}
    \subfloat{\includegraphics[width=0.32\linewidth]{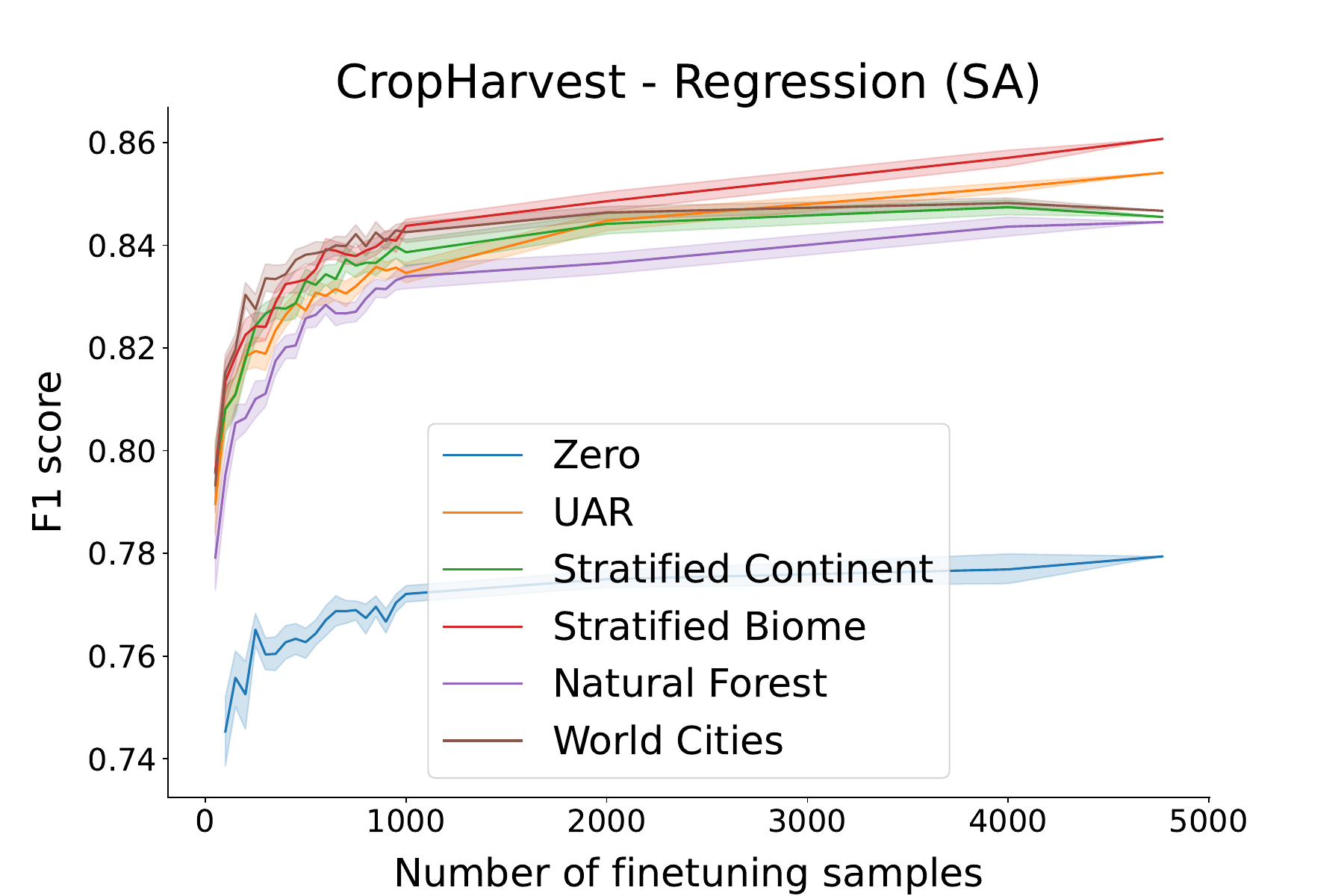}}
    \caption{Continent-wise results for CropHarvest task with Regression}
    \label{fig:result_cropharvest_regression}
\end{figure*}

\begin{figure*}[t]
    \centering
    \subfloat{\includegraphics[width=0.32\linewidth]{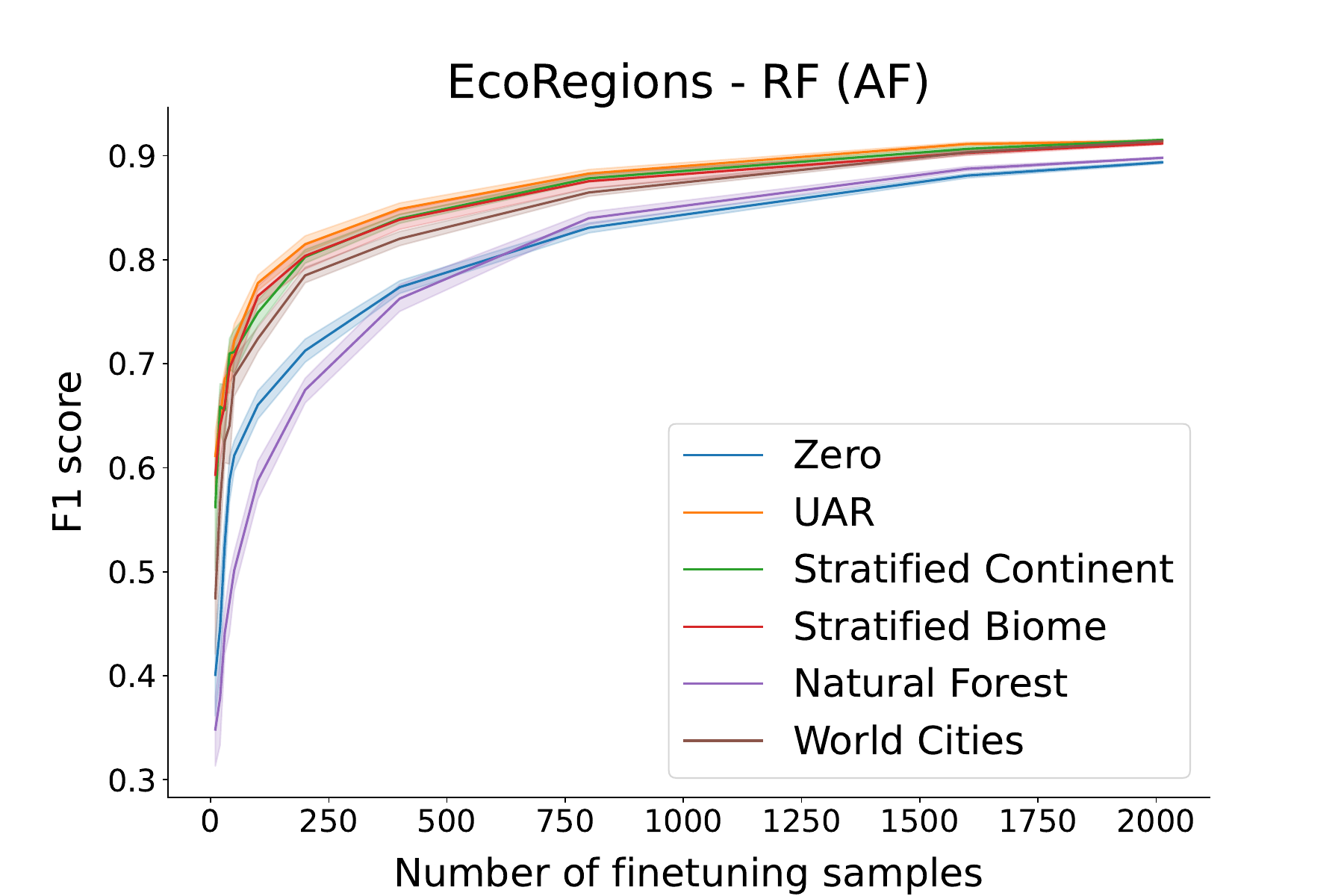}}
    \subfloat{\includegraphics[width=0.32\linewidth]{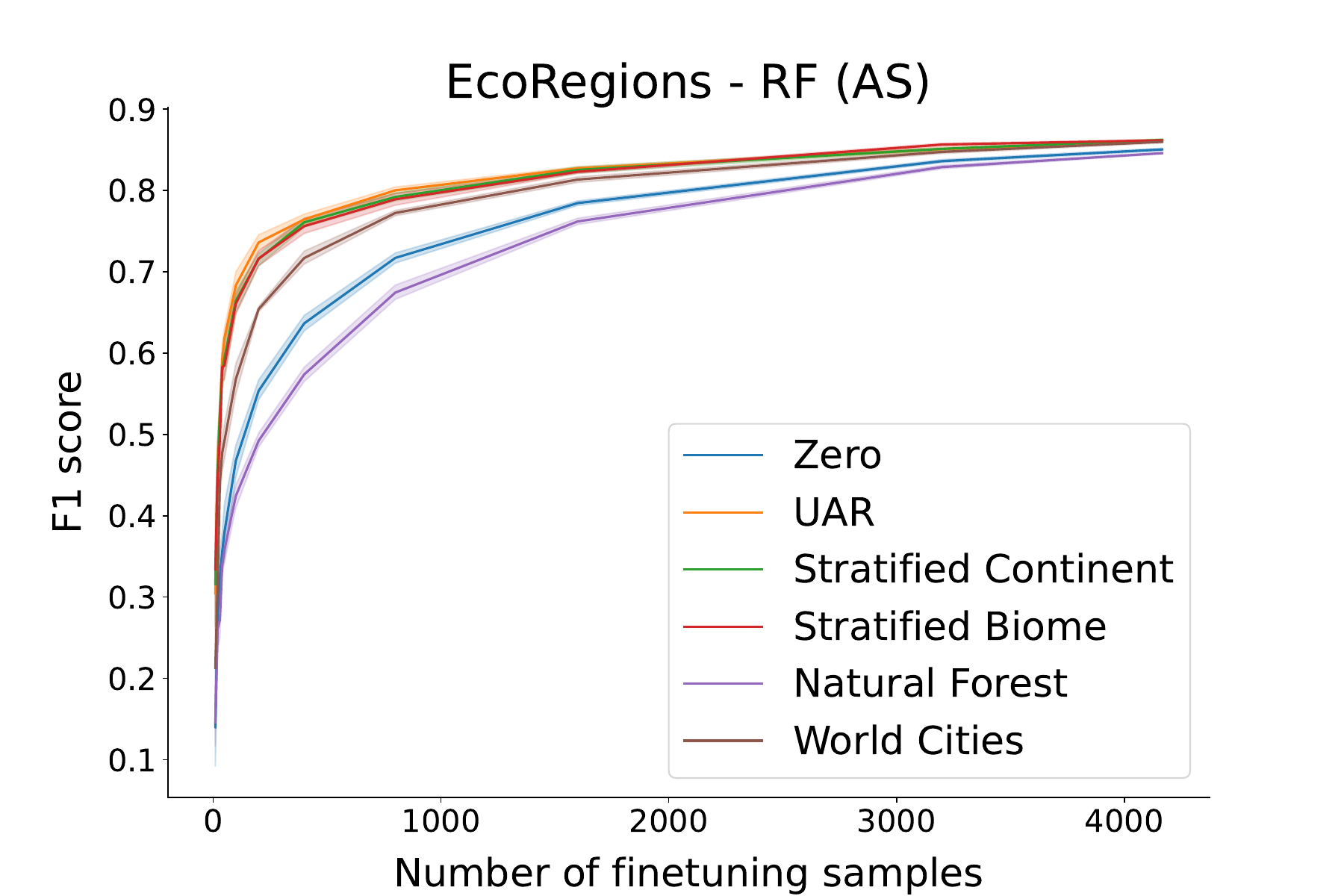}}
    \subfloat{\includegraphics[width=0.32\linewidth]{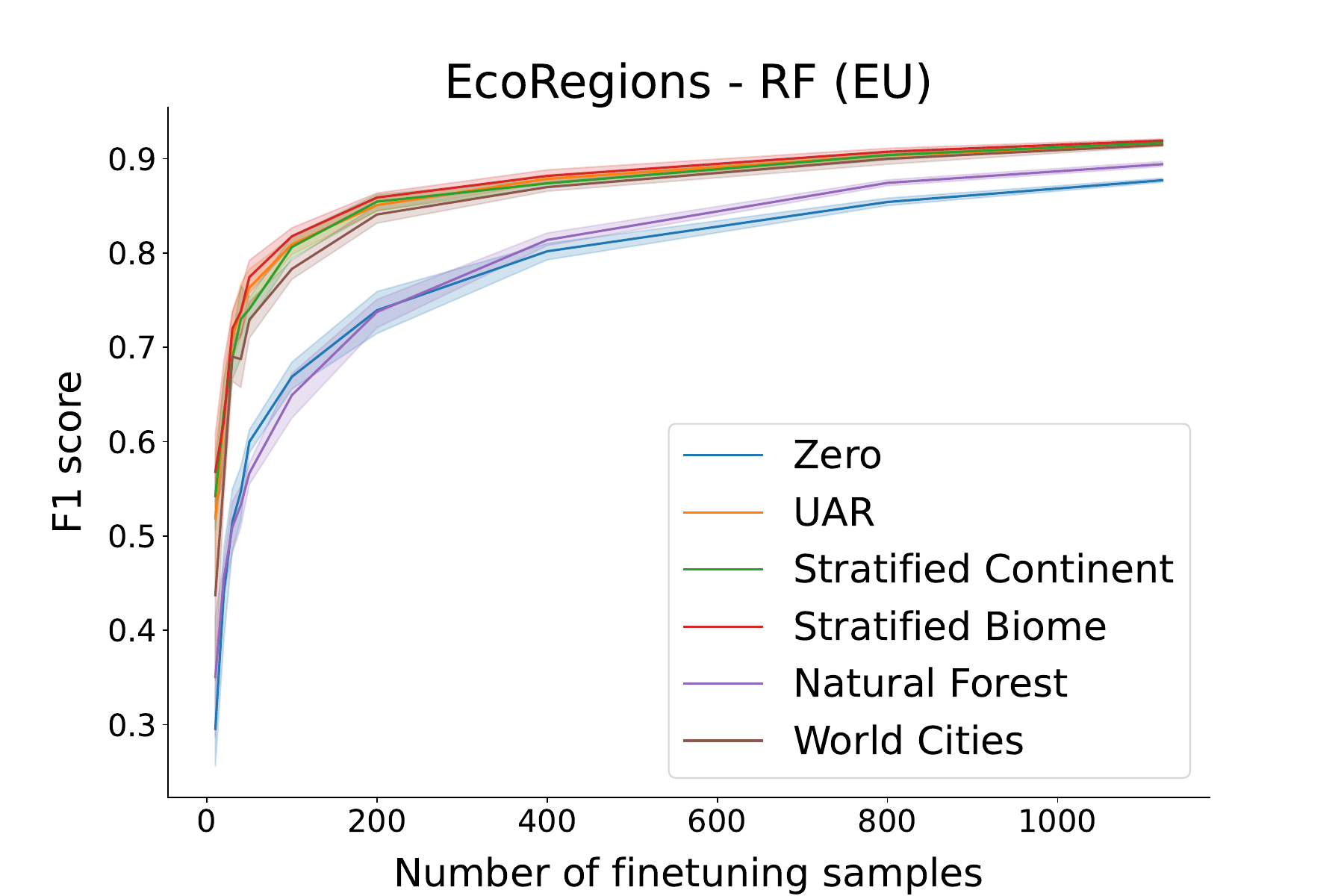}}\\
    \subfloat{\includegraphics[width=0.32\linewidth]{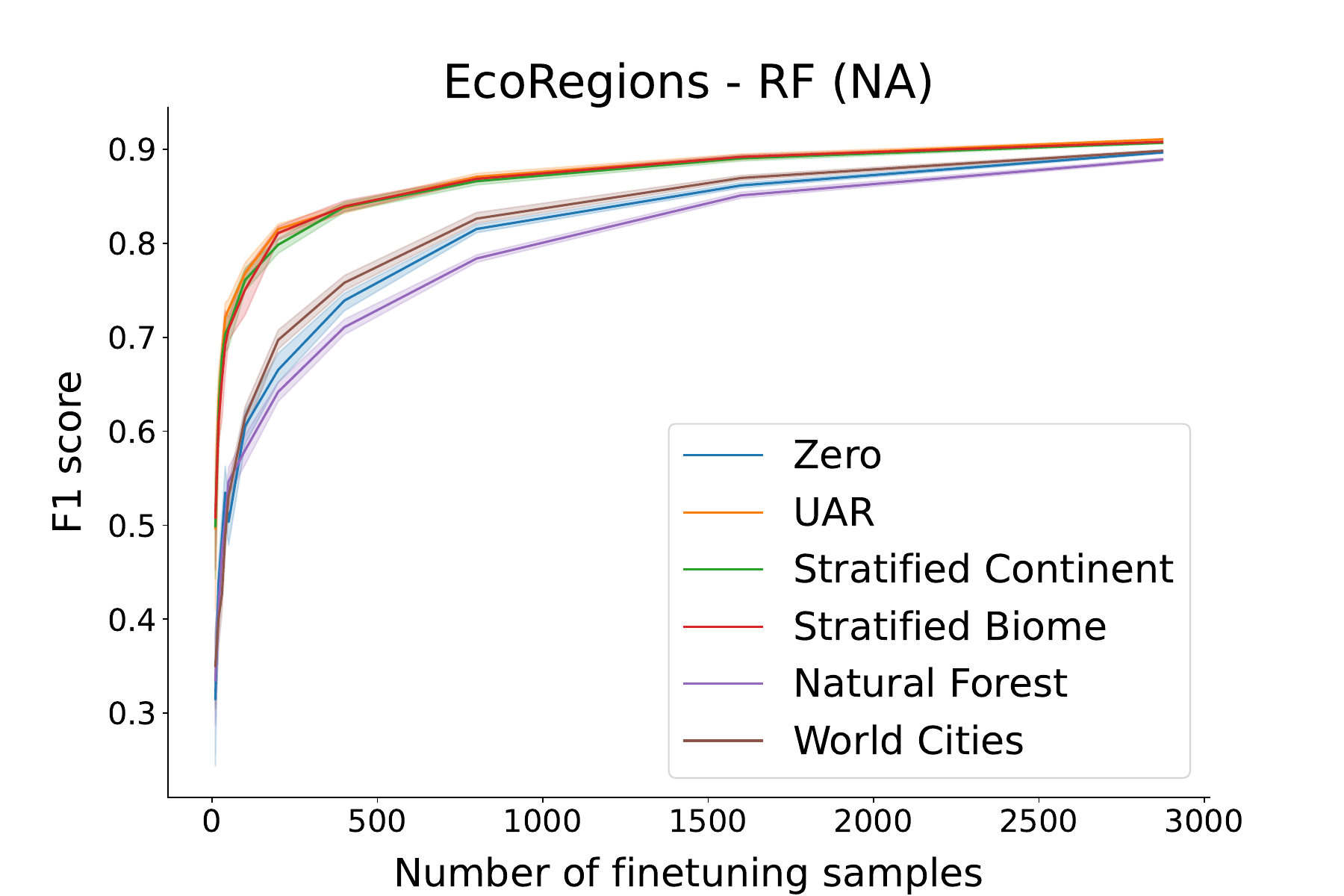}}
    \subfloat{\includegraphics[width=0.32\linewidth]{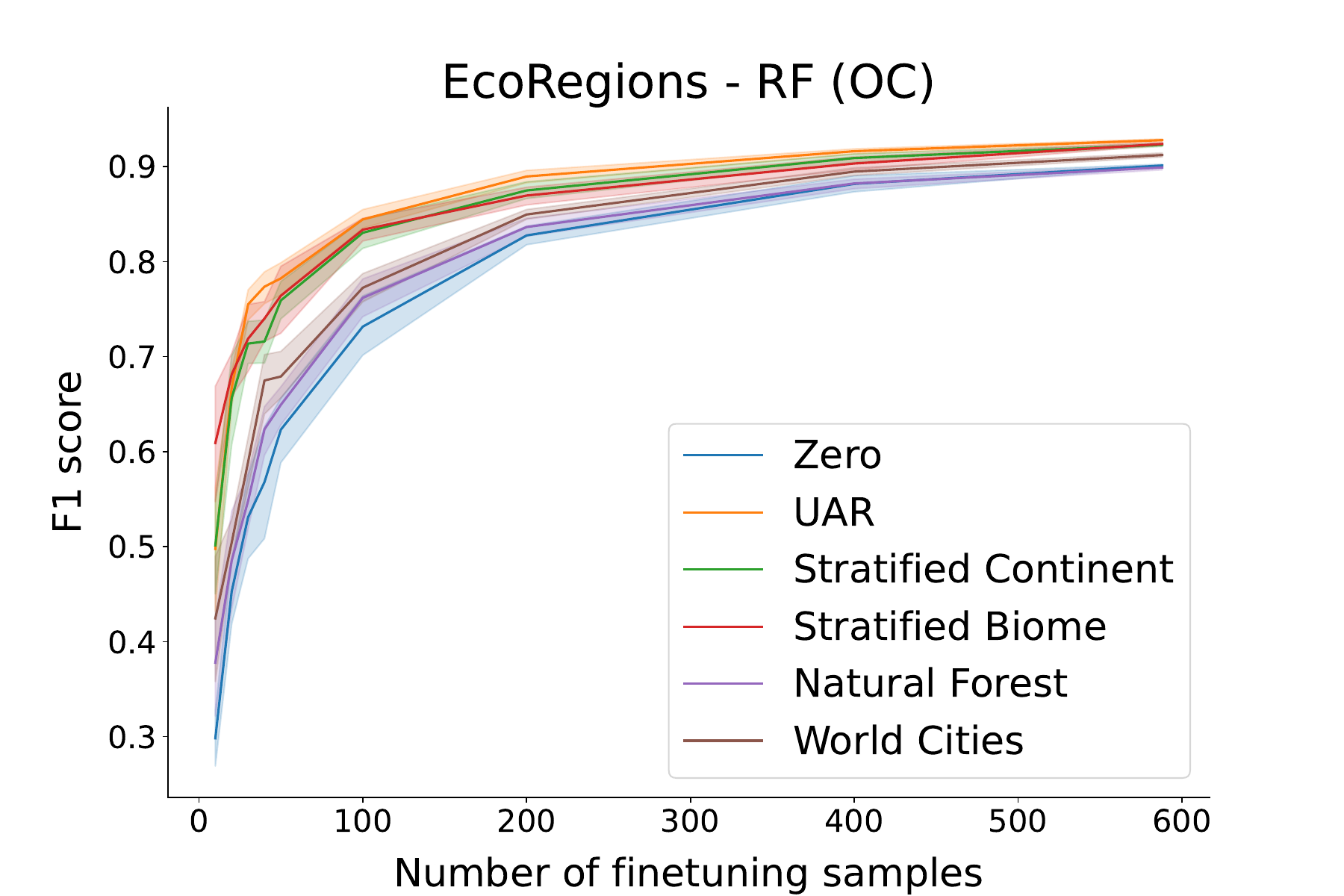}}
    \subfloat{\includegraphics[width=0.32\linewidth]{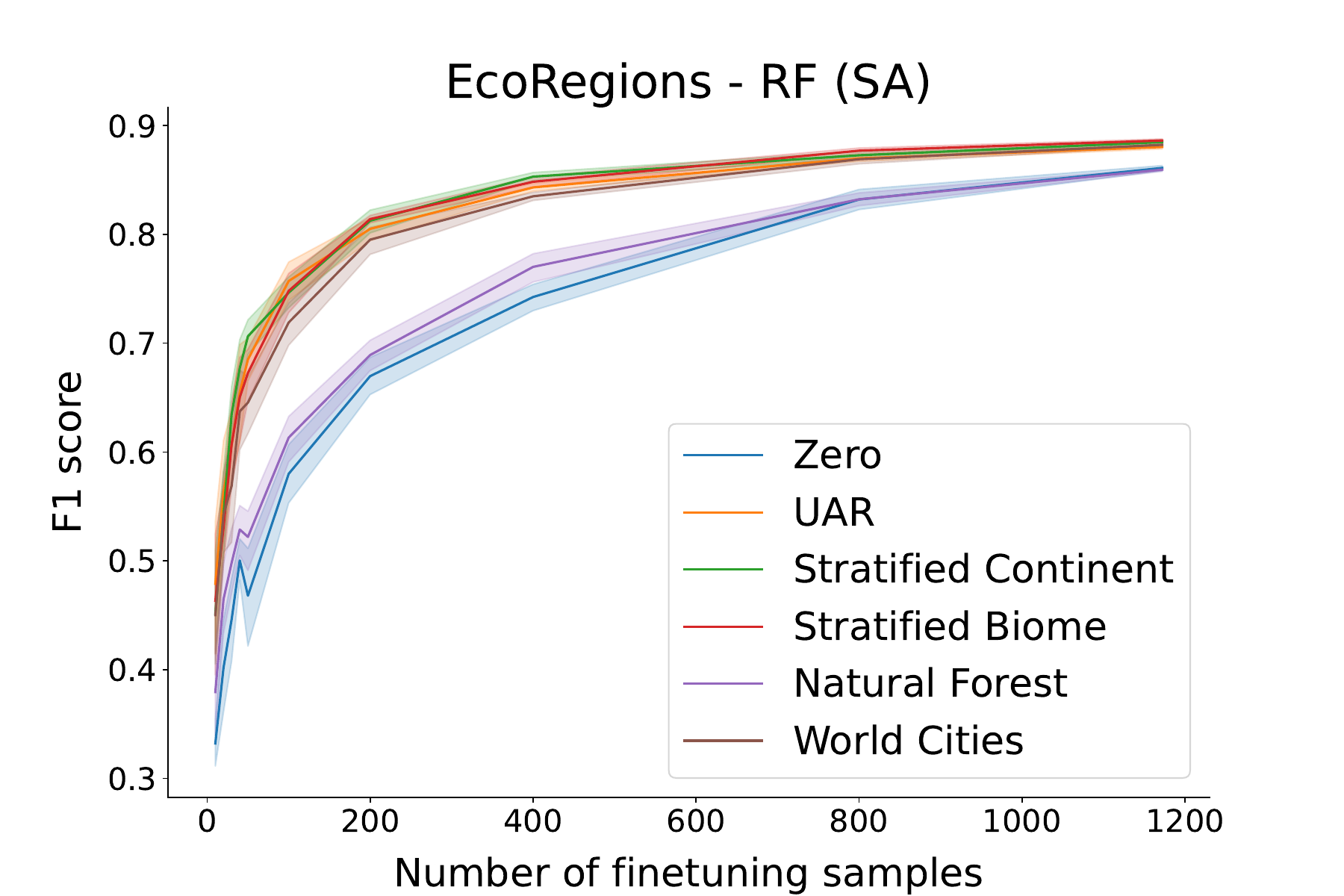}}
    \caption{Continent-wise results for EcoRegions task with Random Forest}
    \label{fig:result_ecoregions_rf}
\end{figure*}

\begin{figure*}[t]
    \centering
    \subfloat{\includegraphics[width=0.32\linewidth]{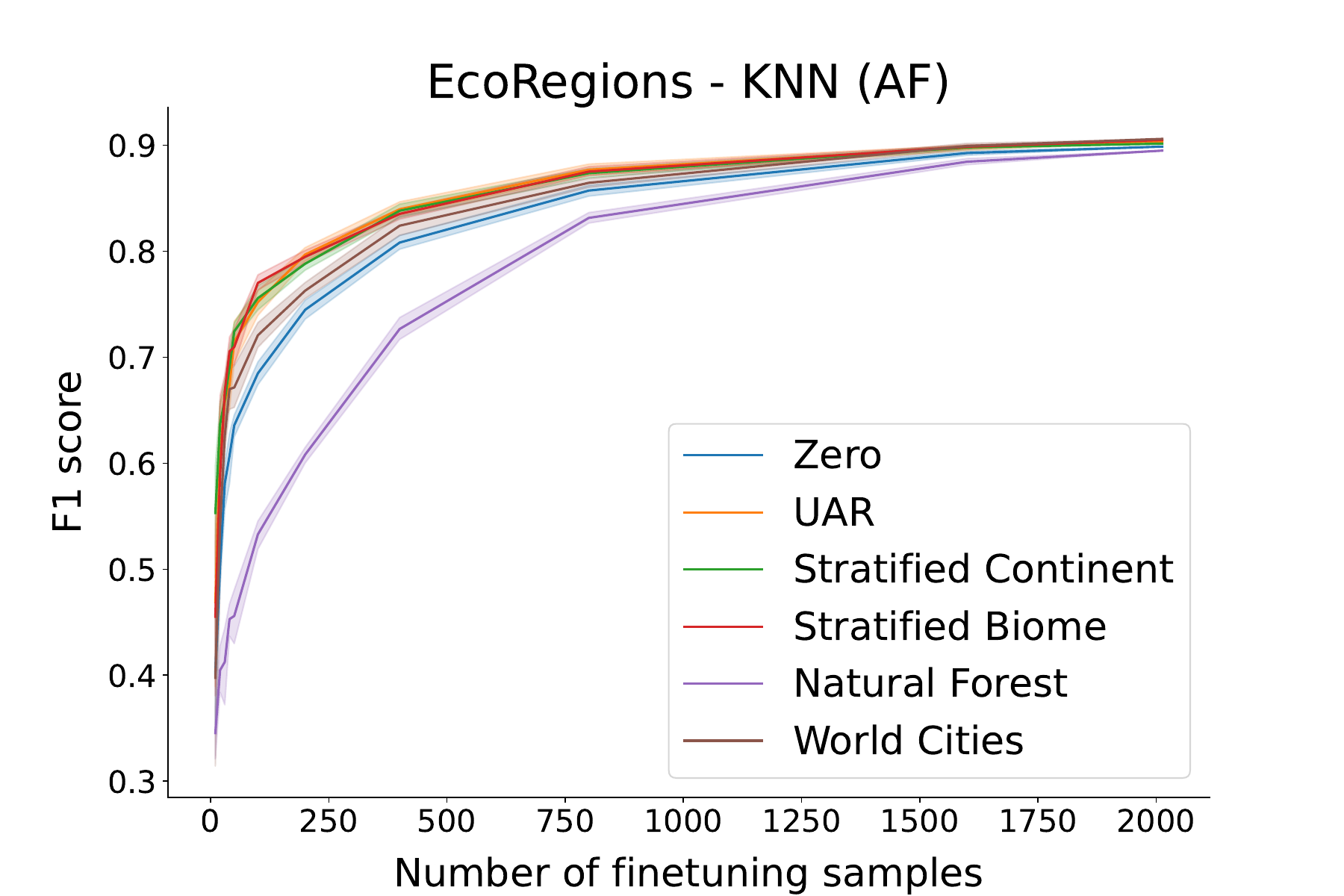}}
    \subfloat{\includegraphics[width=0.32\linewidth]{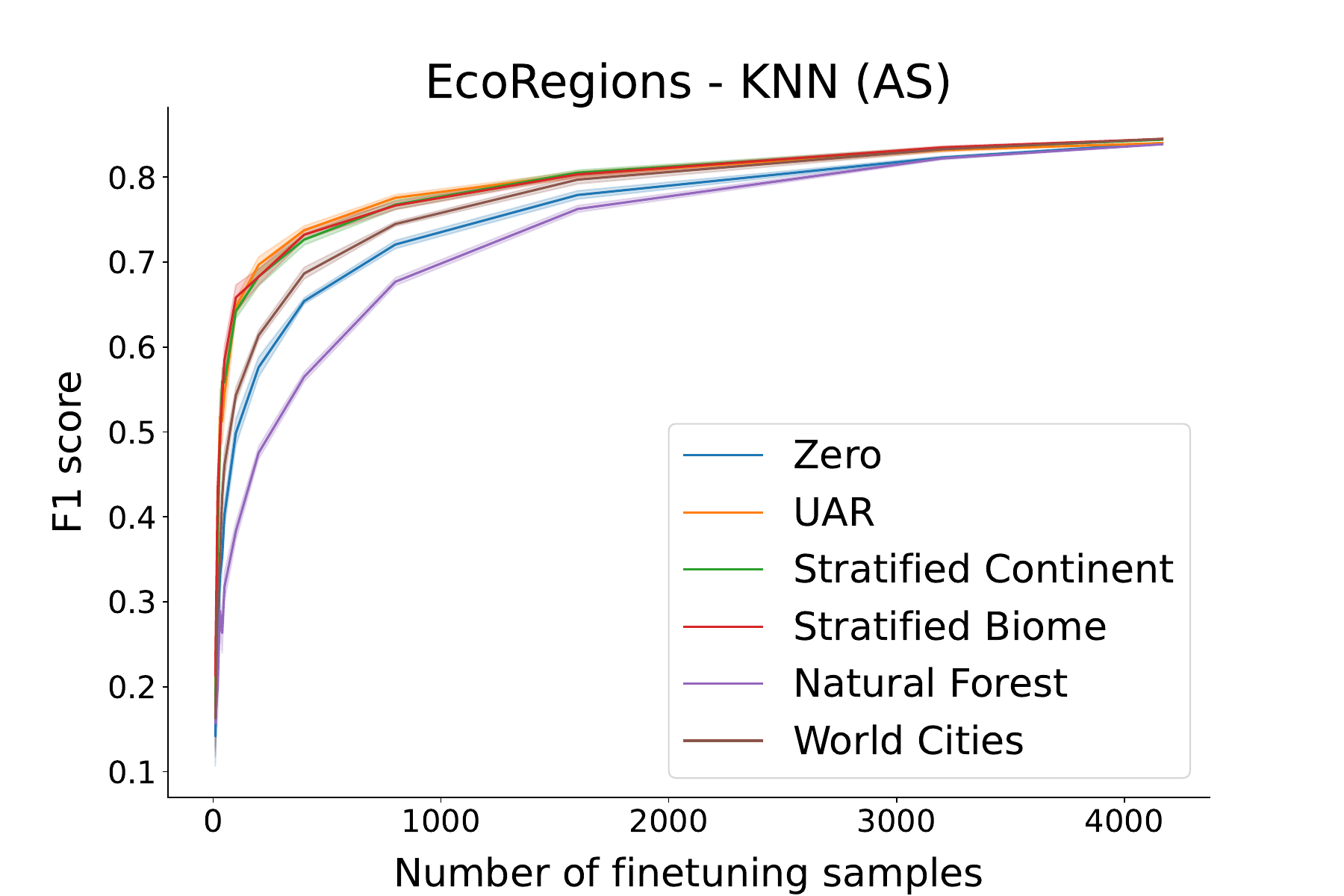}}
    \subfloat{\includegraphics[width=0.32\linewidth]{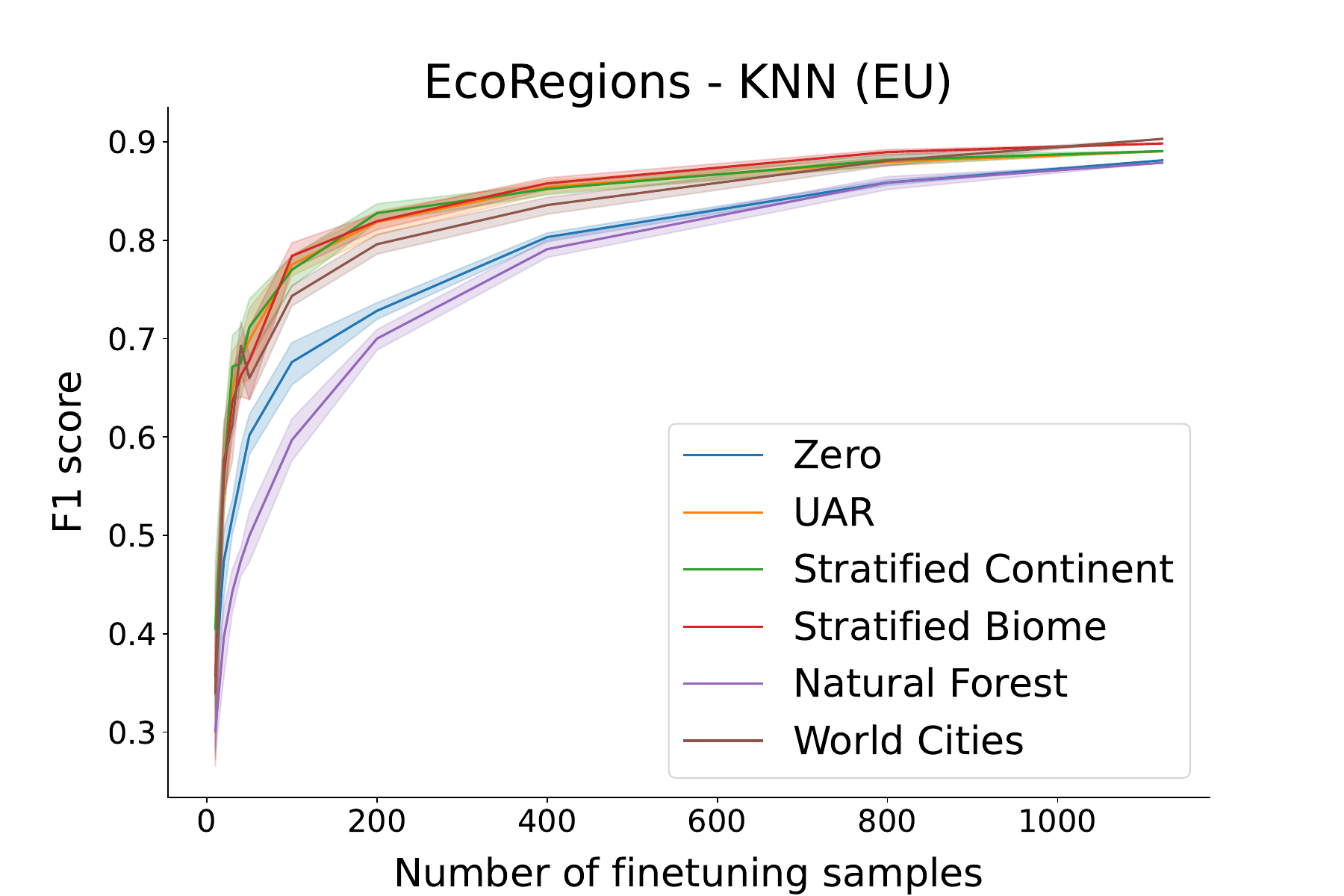}}\\
    \subfloat{\includegraphics[width=0.32\linewidth]{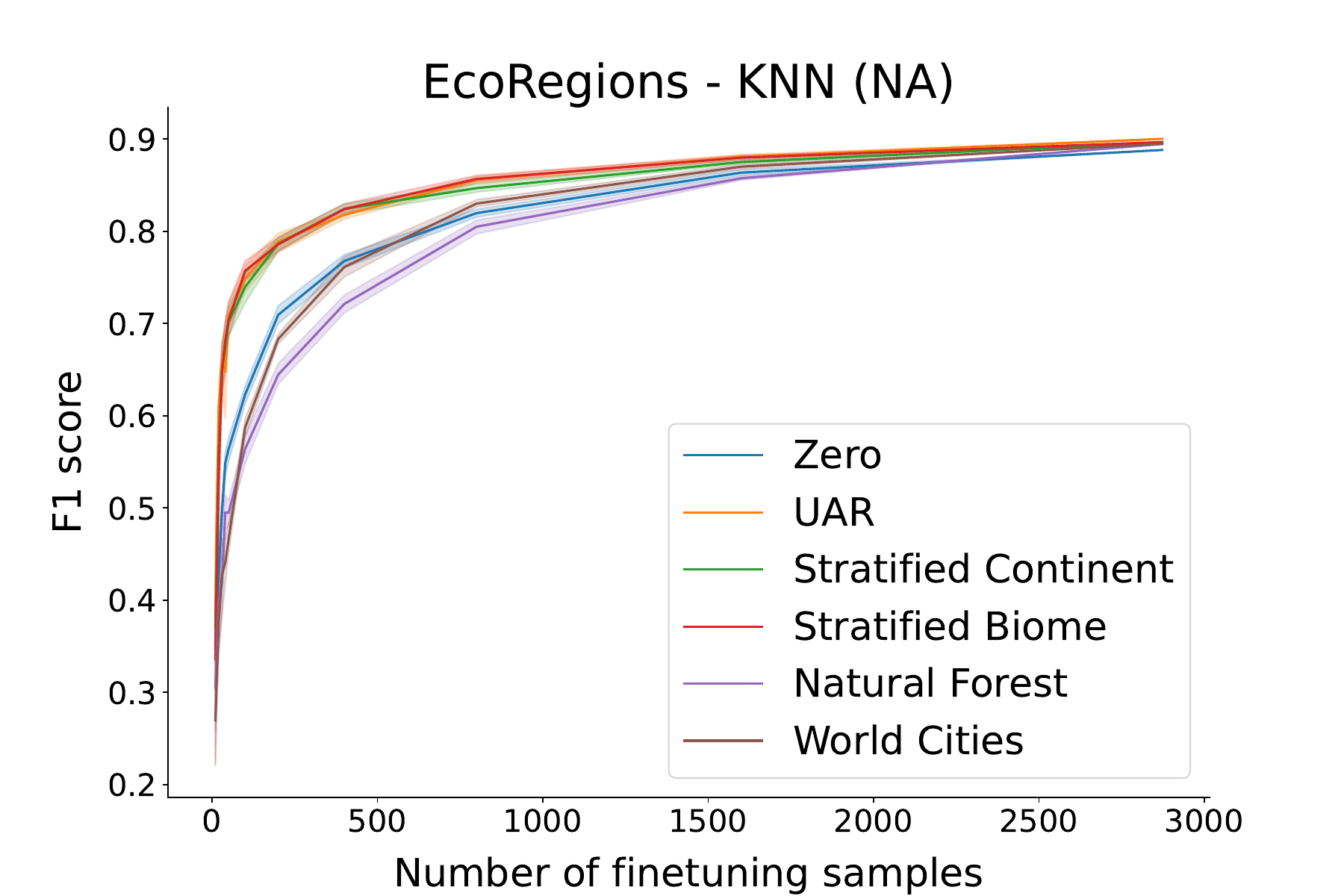}}
    \subfloat{\includegraphics[width=0.32\linewidth]{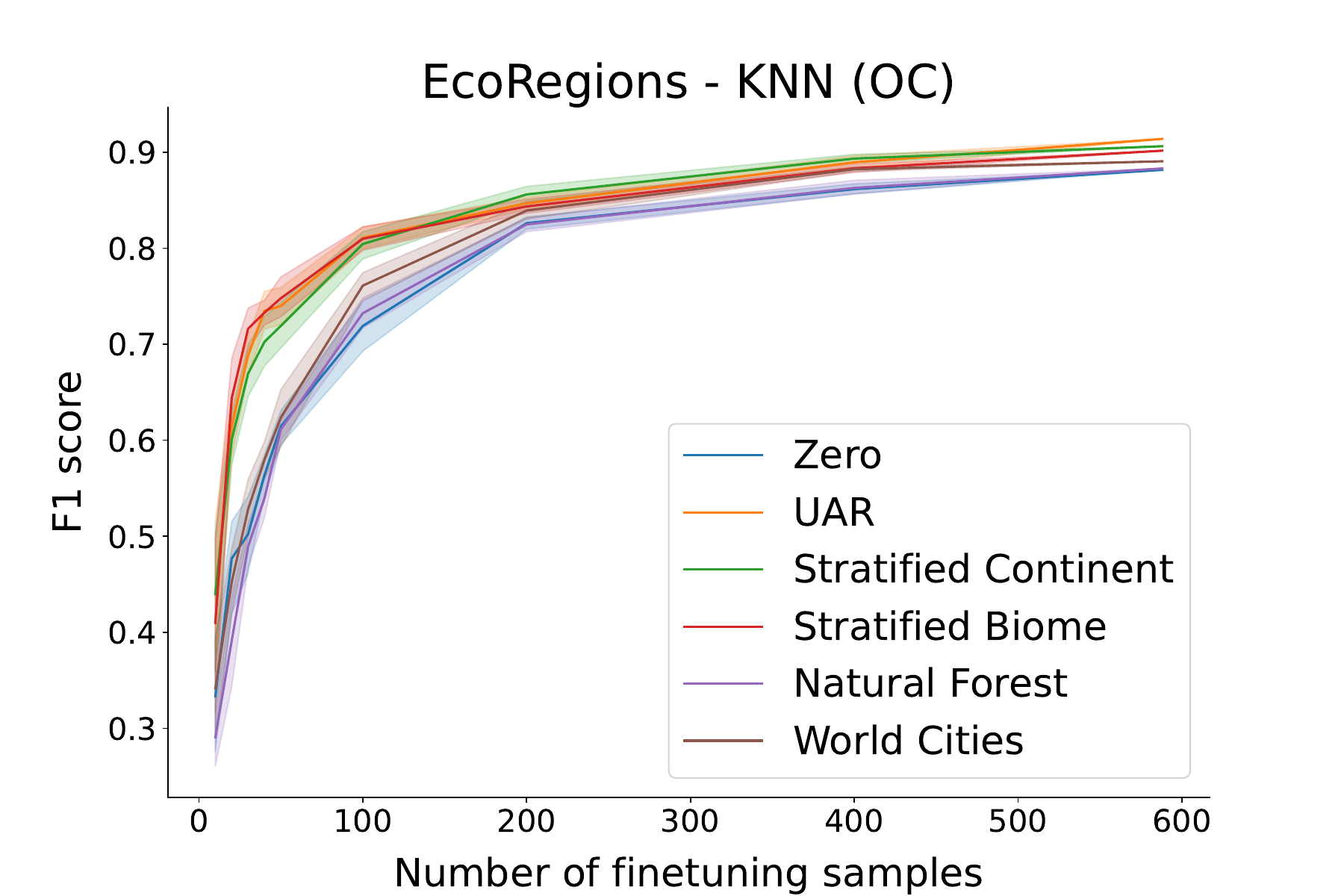}}
    \subfloat{\includegraphics[width=0.32\linewidth]{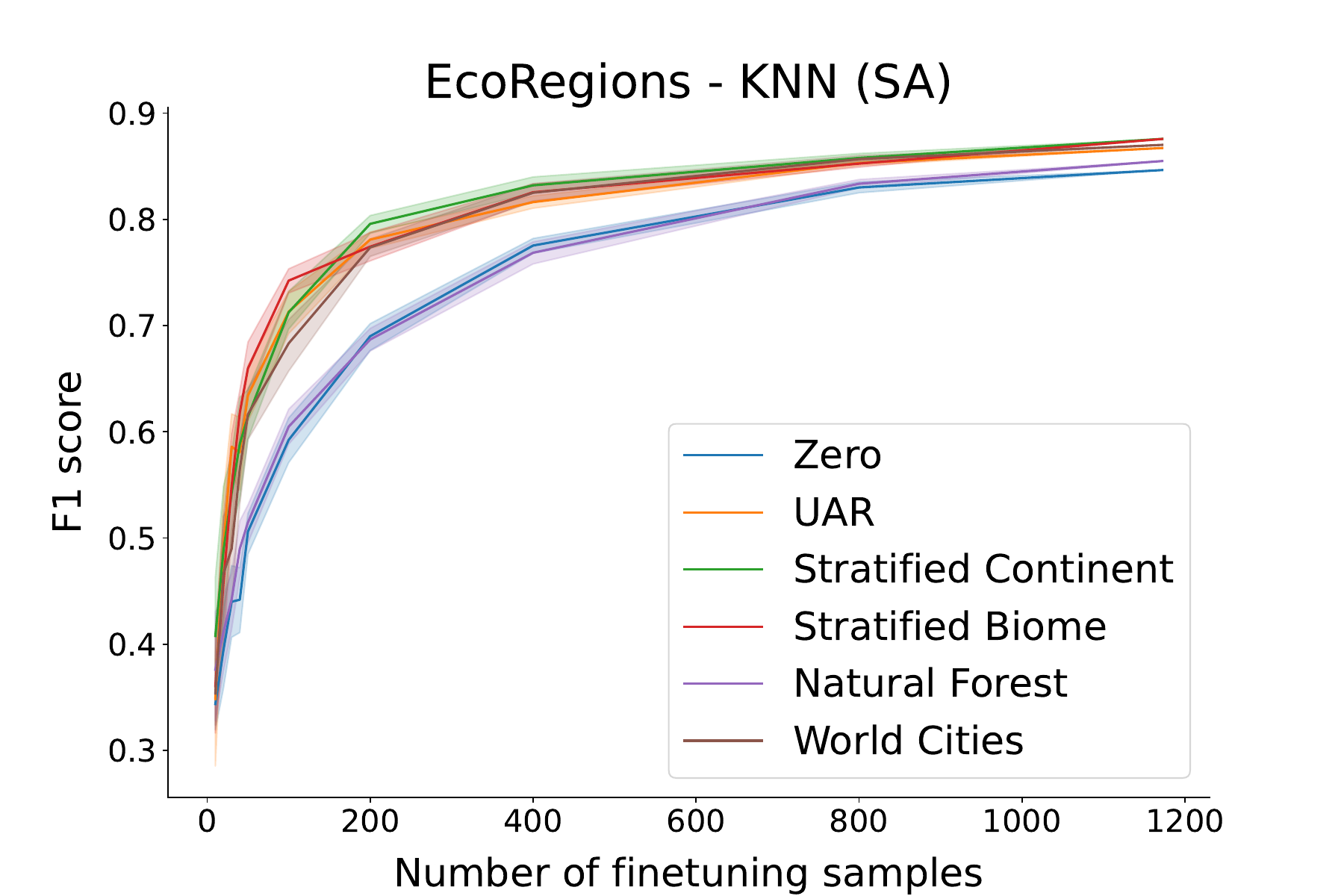}}
    \caption{Continent-wise results for EcoRegions task with KNN}
    \label{fig:result_ecoregions_knn}
\end{figure*}

\begin{figure*}[t]
    \centering
    \subfloat{\includegraphics[width=0.32\linewidth]{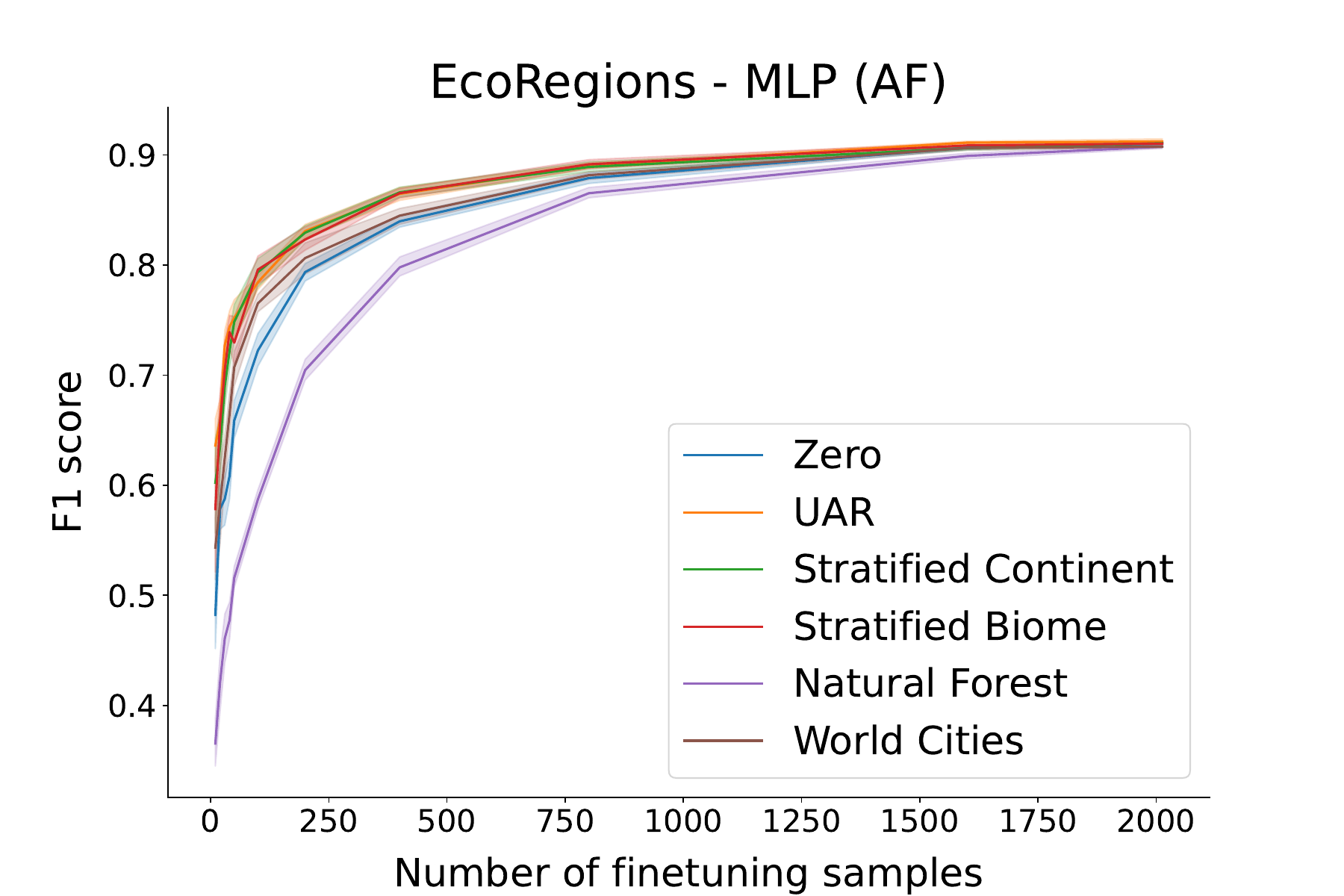}}
    \subfloat{\includegraphics[width=0.32\linewidth]{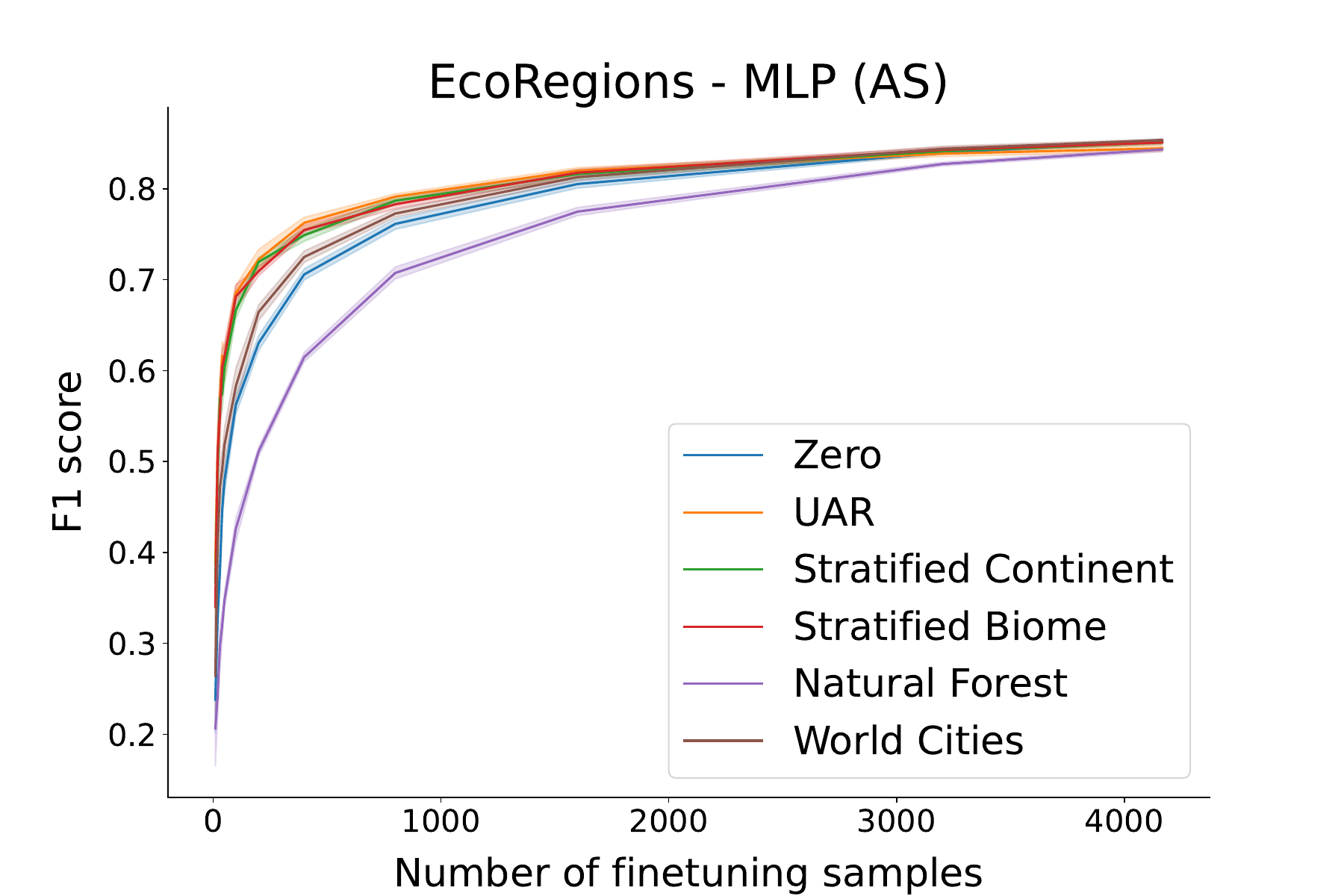}}
    \subfloat{\includegraphics[width=0.32\linewidth]{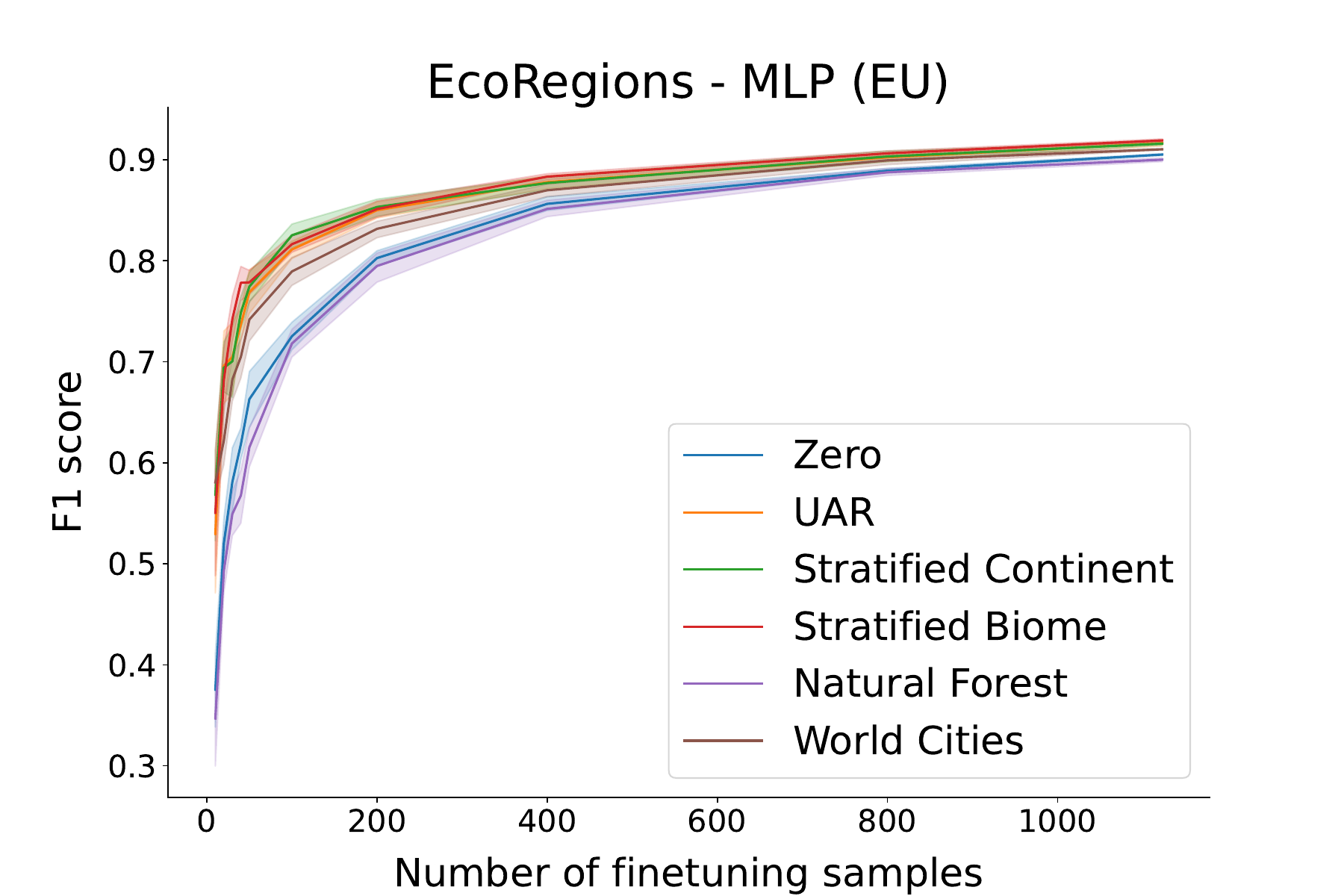}}\\
    \subfloat{\includegraphics[width=0.32\linewidth]{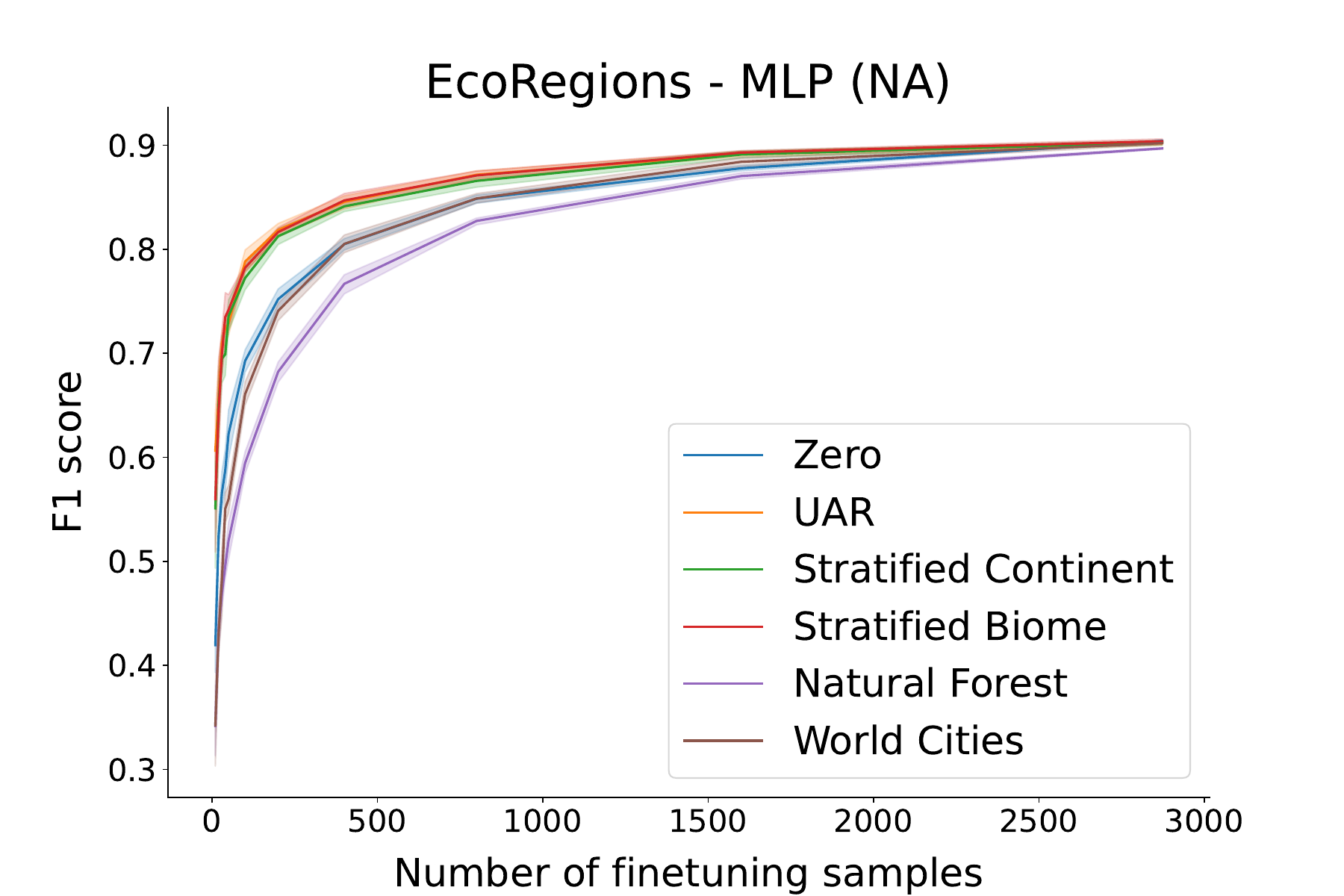}}
    \subfloat{\includegraphics[width=0.32\linewidth]{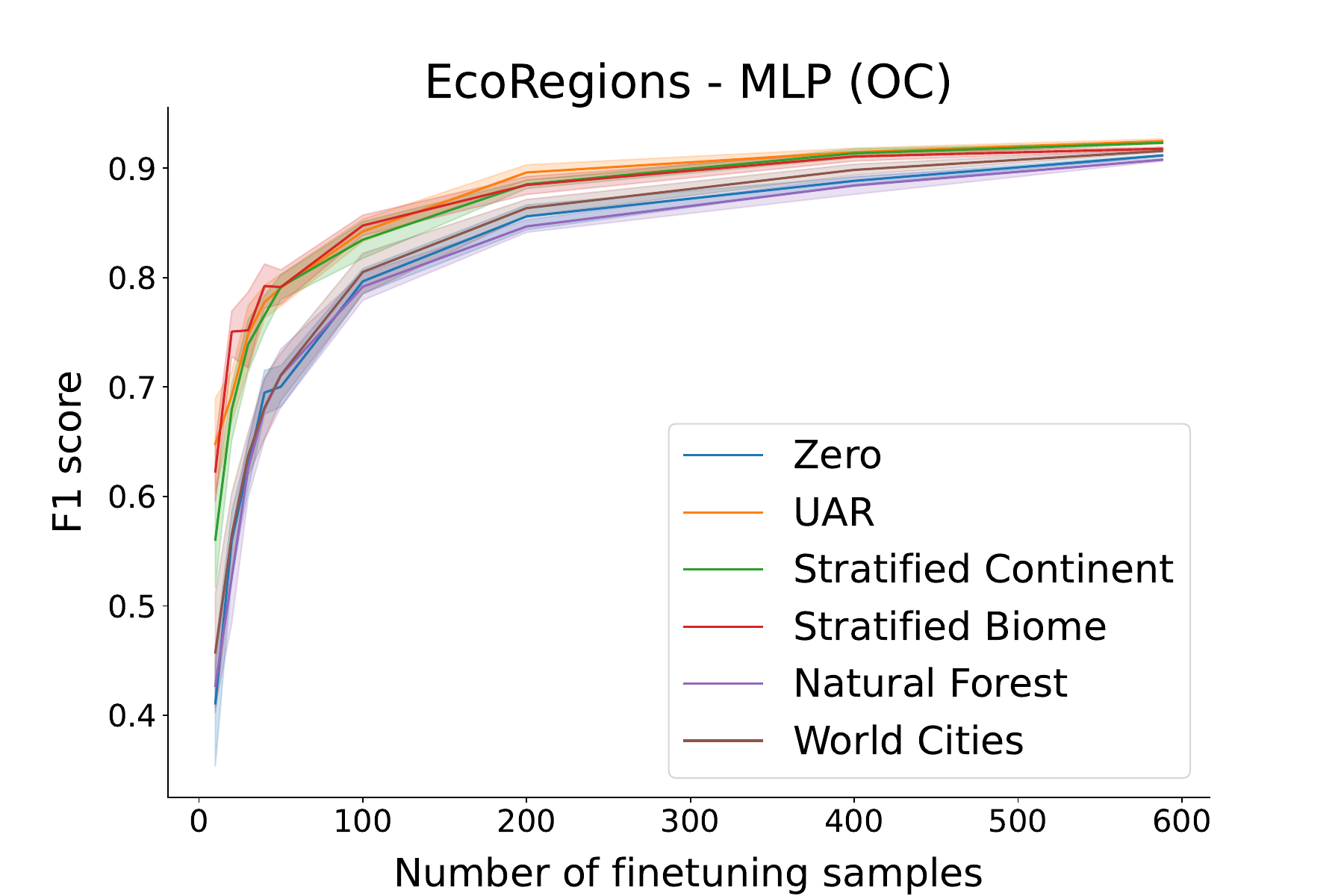}}
    \subfloat{\includegraphics[width=0.32\linewidth]{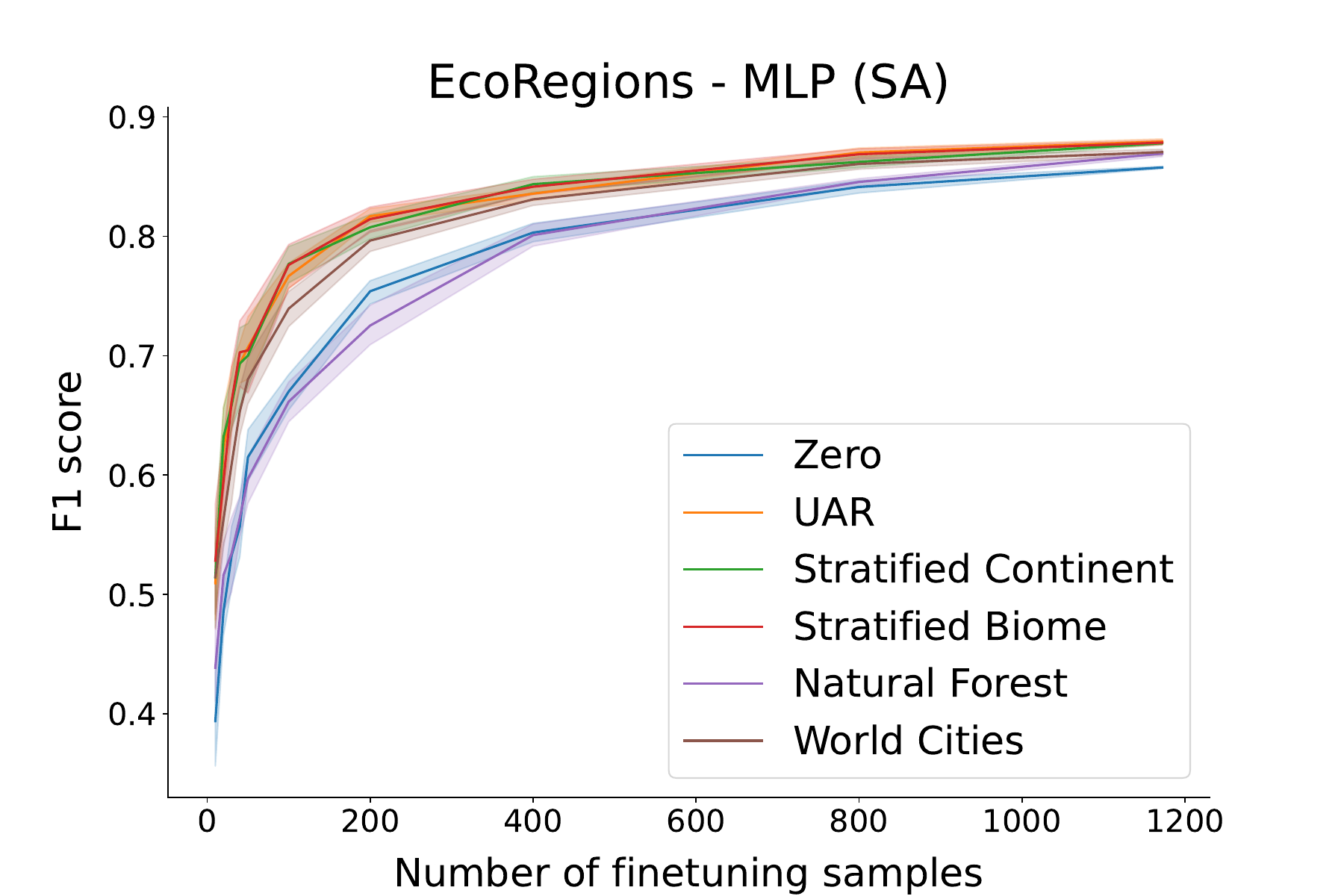}}
    \caption{Continent-wise results for EcoRegions task with MLP}
    \label{fig:result_ecoregions_mlp}
\end{figure*}

Figure \ref{fig:result_cropharvest_rf}, \ref{fig:result_cropharvest_knn}, and \ref{fig:result_cropharvest_regression} present line plots of the F1 score with confidence intervals for CropHarvest, finetuned on RF, KNN, and Regression, respectively. Similarly, results for EcoRegions finetuned using RF, KNN, and MLP are shown in Figure \ref{fig:result_ecoregions_rf}, \ref{fig:result_ecoregions_knn}, and \ref{fig:result_ecoregions_mlp}, respectively. Across all these line plots, we observe that performance improves for all continents as the number of samples increases. Moreover, findings reported in the main paper remain consistent: the impact of differences in data composition diminishes as the sample size grows.


\section{Limitations and Future Work}
\label{appendix_5:future_work}

This study focuses primarily on providing preliminary insights into the impact of pre-training data distribution on the performance of GFMs. There is significant room for extending this research that helps in development of robust GFMs. Below, we outline key areas for future exploration:

Our analysis includes two GFMs and a single downstream task per GFM. Future work can be expanded to evaluate a broader range of GFMs and multiple downstream tasks. This will help to strengthen the findings reported in this research and provide guidelines for pre-training data sampling techniques tailored to diverse EO applications.

In terms of describing geographic diversity, this study selected groupings based on continents, biomes, cities, and natural forests. However, there are many other factors such as countries, hemispheres, climate zones, land cover/land use, etc can also be used which raises an important question: \textit{What groupings should we define pre-training data diversity with respect to?}


Our experiments are conducted at the continent-level due to the infeasibility of country-level finetuning and evaluation (details in Appendix \ref{appendix_3:experiment_details}). Future work can consider downstream tasks with well-distributed data across countries, enabling country-level evaluations and exploring the pre-training data distributions best suited for such granular analyses.

Although this work focuses on balanced and stratified sampling methods for continent and biome grouping, future studies can explore scenarios where intentionally (it can be somewhat or totally) biased sampling techniques can enhance the performance for specific regional tasks. This will help to understand the trade-offs between global coverage and regional specialization in GFM development.

While this paper focuses on the impact of geographic diversity of pre-training data, the influence of data quantity is another crucial factor. Future research can investigate the relationship between the volume of pre-training data and model performance, exploring whether increasing the amount of data leads to proportional improvements in downstream task performance.

By addressing these gaps, the GeoAI community can develop more effective GFMs and establish best practices for pre-training data selection, ultimately advancing machine learning solutions for Earth observation tasks.

\end{document}

%% file: tables/result_mean_2_std_2.tex
\begin{table*}[!ht]
\setlength\tabcolsep{4.0pt}
\setlength{\belowcaptionskip}{-10pt}
\centering
\caption{Average F1 score and standard deviation (over $50$ runs) for the CropHarvest and EcoRegions when Presto and SatCLIP are finetuned using \underline{Random Forest} (RF) on $100$ training samples, respectively. Here, \tcbox[colback=light_red]{highlighted} results indicate scores that are at least $2 \%$ lower than UAR. (AF: Africa, AS: Asia, EU: Europe, NA: North America, OC: Oceania, SA: South America)}
\footnotesize
\resizebox{0.98\linewidth}{!}{
\begin{tabular}{c|cccccc|cccccc}
\toprule[1.5pt]
\multirow{2}{*}{\textbf{\begin{tabular}[c]{@{}c@{}}Data\\Composition\end{tabular}}} &
  \multicolumn{6}{c|}{\textbf{CropHarvest}} &
  \multicolumn{6}{c}{\textbf{EcoRegion}} \\ \cmidrule{2-13} 
 &
  \multicolumn{1}{c}{\textbf{AF}} &
  \multicolumn{1}{c}{\textbf{AS}} &
  \multicolumn{1}{c}{\textbf{EU}} &
  \multicolumn{1}{c}{\textbf{NA}} &
  \multicolumn{1}{c}{\textbf{OC}} &
  \multicolumn{1}{c|}{\textbf{SA}} &

  \multicolumn{1}{c}{\textbf{AF}} &
  \multicolumn{1}{c}{\textbf{AS}} &
  \multicolumn{1}{c}{\textbf{EU}} &
  \multicolumn{1}{c}{\textbf{NA}} &
  \multicolumn{1}{c}{\textbf{OC}} &
  \multicolumn{1}{c}{\textbf{SA}} \\ \midrule[1pt]

Zero &
  \multicolumn{1}{c}{\tcbox[colback=light_red]{0.68 $\pm$ 0.02}} &
  \multicolumn{1}{c}{\tcbox[colback=light_red]{0.76 $\pm$ 0.03}} &
  \multicolumn{1}{c}{\tcbox[colback=light_red]{0.74 $\pm$ 0.02}} &
  \multicolumn{1}{c}{0.79 $\pm$ 0.02} &
  \multicolumn{1}{c}{\tcbox[colback=light_red]{0.73 $\pm$ 0.03}} &
  \multicolumn{1}{c|}{\tcbox[colback=light_red]{0.75 $\pm$ 0.02}} &

  \multicolumn{1}{c}{\tcbox[colback=light_red]{0.66 $\pm$ 0.02}} &
  \multicolumn{1}{c}{\tcbox[colback=light_red]{0.47 $\pm$ 0.03}} &
  \multicolumn{1}{c}{\tcbox[colback=light_red]{0.67 $\pm$ 0.02}} &
  \multicolumn{1}{c}{\tcbox[colback=light_red]{0.61 $\pm$ 0.03}} &
  \multicolumn{1}{c}{\tcbox[colback=light_red]{0.73 $\pm$ 0.05}} &
  \multicolumn{1}{c}{\tcbox[colback=light_red]{0.58 $\pm$ 0.04}} \\ \midrule

UAR &
  \multicolumn{1}{c}{0.71 $\pm$ 0.02} &
  \multicolumn{1}{c}{0.78 $\pm$ 0.03} &
  \multicolumn{1}{c}{0.76 $\pm$ 0.02} &
  \multicolumn{1}{c}{0.80 $\pm$ 0.03} &
  \multicolumn{1}{c}{0.75 $\pm$ 0.03} &
  \multicolumn{1}{c|}{0.79 $\pm$ 0.02} &

  \multicolumn{1}{c}{0.78 $\pm$ 0.01} &
  \multicolumn{1}{c}{0.68 $\pm$ 0.03} &
  \multicolumn{1}{c}{0.81 $\pm$ 0.02} &
  \multicolumn{1}{c}{0.77 $\pm$ 0.02} &
  \multicolumn{1}{c}{0.84 $\pm$ 0.02} &
  \multicolumn{1}{c}{0.76 $\pm$ 0.03} \\ \midrule

Stratified Continent &
  \multicolumn{1}{c}{0.71 $\pm$ 0.02} &
  \multicolumn{1}{c}{0.78 $\pm$ 0.03} &
  \multicolumn{1}{c}{0.76 $\pm$ 0.03} &
  \multicolumn{1}{c}{0.81 $\pm$ 0.03} &
  \multicolumn{1}{c}{0.75 $\pm$ 0.02} &
  \multicolumn{1}{c|}{0.80 $\pm$ 0.02} &

  \multicolumn{1}{c}{0.75 $\pm$ 0.02} &
  \multicolumn{1}{c}{0.66 $\pm$ 0.02} &
  \multicolumn{1}{c}{0.81 $\pm$ 0.02} &
  \multicolumn{1}{c}{0.76 $\pm$ 0.02} &
  \multicolumn{1}{c}{0.83 $\pm$ 0.03} &
  \multicolumn{1}{c}{0.75 $\pm$ 0.02} \\ \midrule

Stratified Biome &
  \multicolumn{1}{c}{0.71 $\pm$ 0.02} &
  \multicolumn{1}{c}{0.78 $\pm$ 0.03} &
  \multicolumn{1}{c}{0.75 $\pm$ 0.02} &
  \multicolumn{1}{c}{0.81 $\pm$ 0.02} &
  \multicolumn{1}{c}{0.76 $\pm$ 0.03} &
  \multicolumn{1}{c|}{0.80 $\pm$ 0.02} &

  \multicolumn{1}{c}{0.77 $\pm$ 0.02} &
  \multicolumn{1}{c}{0.66 $\pm$ 0.02} &
  \multicolumn{1}{c}{0.82 $\pm$ 0.02} &
  \multicolumn{1}{c}{0.75 $\pm$ 0.04} &
  \multicolumn{1}{c}{0.83 $\pm$ 0.02} &
  \multicolumn{1}{c}{0.75 $\pm$ 0.03} \\ \midrule

Natural Forest &
  \multicolumn{1}{c}{\tcbox[colback=light_red]{0.67 $\pm$ 0.02}} &
  \multicolumn{1}{c}{\tcbox[colback=light_red]{0.75 $\pm$ 0.03}} &
  \multicolumn{1}{c}{\tcbox[colback=light_red]{0.73 $\pm$ 0.02}} &
  \multicolumn{1}{c}{\tcbox[colback=light_red]{0.77 $\pm$ 0.03}} &
  \multicolumn{1}{c}{0.75 $\pm$ 0.03} &
  \multicolumn{1}{c|}{0.80 $\pm$ 0.03} &

  \multicolumn{1}{c}{\tcbox[colback=light_red]{0.59 $\pm$ 0.03}} &
  \multicolumn{1}{c}{\tcbox[colback=light_red]{0.42 $\pm$ 0.02}} &
  \multicolumn{1}{c}{\tcbox[colback=light_red]{0.65 $\pm$ 0.04}} &
  \multicolumn{1}{c}{\tcbox[colback=light_red]{0.58 $\pm$ 0.03}} &
  \multicolumn{1}{c}{\tcbox[colback=light_red]{0.76 $\pm$ 0.03}} &
  \multicolumn{1}{c}{\tcbox[colback=light_red]{0.61 $\pm$ 0.04}} \\ \midrule

World Cities &
  \multicolumn{1}{c}{0.72 $\pm$ 0.02} &
  \multicolumn{1}{c}{0.78 $\pm$ 0.03} &
  \multicolumn{1}{c}{0.76 $\pm$ 0.03} &
  \multicolumn{1}{c}{0.81 $\pm$ 0.03} &
  \multicolumn{1}{c}{0.75 $\pm$ 0.02} &
  \multicolumn{1}{c|}{0.79 $\pm$ 0.02} &

  \multicolumn{1}{c}{\tcbox[colback=light_red]{0.72 $\pm$ 0.02}} &
  \multicolumn{1}{c}{\tcbox[colback=light_red]{0.57 $\pm$ 0.03}} &
  \multicolumn{1}{c}{\tcbox[colback=light_red]{0.78 $\pm$ 0.02}} &
  \multicolumn{1}{c}{\tcbox[colback=light_red]{0.61 $\pm$ 0.02}} &
  \multicolumn{1}{c}{\tcbox[colback=light_red]{0.77 $\pm$ 0.03}} &
  \multicolumn{1}{c}{\tcbox[colback=light_red]{0.72 $\pm$ 0.03}} \\ \bottomrule
\end{tabular}
}
\label{tab:results}
\end{table*}